\documentclass[a4paper, 10pt, conference]{ieeeconf}      

\usepackage[utf8]{inputenc}

\IEEEoverridecommandlockouts                              

\usepackage[caption=false,font=footnotesize]{subfig}

\usepackage{graphics} 
\usepackage{cite}
\usepackage{todonotes}
\usepackage{siunitx}

\usepackage{pgfplots}
\pgfplotsset{compat=newest}
\pgfplotsset{plot coordinates/math parser=false}
\newlength\figureheight
\newlength\figurewidth

\usepackage{tabularx}
\usepackage{multirow}
\usepackage{makecell}

\usepackage{scalerel}
\usepackage{tikz}
\usetikzlibrary{svg.path}

\definecolor{orcidlogocol}{HTML}{A6CE39}
\tikzset{
  orcidlogo/.pic={
    \fill[orcidlogocol] svg{M256,128c0,70.7-57.3,128-128,128C57.3,256,0,198.7,0,128C0,57.3,57.3,0,128,0C198.7,0,256,57.3,256,128z};
    \fill[white] svg{M86.3,186.2H70.9V79.1h15.4v48.4V186.2z}
                 svg{M108.9,79.1h41.6c39.6,0,57,28.3,57,53.6c0,27.5-21.5,53.6-56.8,53.6h-41.8V79.1z M124.3,172.4h24.5c34.9,0,42.9-26.5,42.9-39.7c0-21.5-13.7-39.7-43.7-39.7h-23.7V172.4z}
                 svg{M88.7,56.8c0,5.5-4.5,10.1-10.1,10.1c-5.6,0-10.1-4.6-10.1-10.1c0-5.6,4.5-10.1,10.1-10.1C84.2,46.7,88.7,51.3,88.7,56.8z};
  }
}

\newcommand\orcidicon[1]{\textsuperscript{\,\href{https://orcid.org/#1}{\mbox{\scalerel*{%
	\begin{tikzpicture}[yscale=-1,transform shape]
		\pic{orcidlogo};
	\end{tikzpicture}
}{|}}}}%
}

\usepackage{varioref}
\usepackage[hidelinks]{hyperref} 
\usepackage[capitalise]{cleveref}

\newcommand\copyrighttext{%
	\footnotesize \copyright~2019 IEEE. Personal use of this material is permitted. Permission from IEEE must be obtained for all other uses, in any current or future media, including reprinting/republishing this material for advertising or promotional purposes, creating new collective works, for resale or redistribution to servers or lists, or reuse of any copyrighted component of this work in other works. DOI: 10.1109/ITSC.2019.8917048}%
\newcommand\copyrightnotice{%
	\begin{tikzpicture}[remember picture,overlay]%
	\node[anchor=south,yshift=10pt] at (current page.south) {\fbox{\parbox{\dimexpr\textwidth-2cm}{\copyrighttext}}};%
	\end{tikzpicture}%
	\vspace{-10pt}%
}

\title{\LARGE \bf
Fusion of Object Tracking and Dynamic Occupancy Grid Map
}

\author{Nils Rexin\orcidicon{0000-0001-9184-330X}%
	, Marcel Musch\orcidicon{0000-0001-9976-2846} and Klaus Dietmayer
\thanks{All authors are from Institute of Measurement, Control, and Microtechnology, Ulm University, Germany
	{\tt\footnotesize \{firstname.name\}@uni-ulm.de}}%
}

\begin{document}

\maketitle%
\copyrightnotice%
\thispagestyle{empty}
\pagestyle{empty}

\begin{abstract}
Environment modeling in autonomous driving is realized by two fundamental approaches, grid-based and feature-based approach. Both methods interpret the environment differently and show some situation-dependent beneficial realizations. In order to use the advantages of both methods, a combination makes sense. This work presents a fusion, which establishes an association between the representations of environment modeling and then decoupled from this performs a fusion of the information. Thus, there is no need to adapt the environment models. The developed fusion generates new hypotheses, which are closer to reality than a representation alone. 
This algorithm itself does not use object model assumptions, in effect this fusion can be applied to different object hypotheses.
In addition, this combination allows the objects to be tracked over a longer period of time. This is evaluated with a quantitative evaluation on real sequences in real-time.
\end{abstract}


\section{Introduction}
For the planning of movements of an autonomous robot, a modeling of the environment is required in order to determine the corresponding action space for a collision-free execution of a movement. All obstacles in the environment have to be considered. In particular, dynamic objects are of interest because they directly influence the free movement space. There are two fundamentally different approaches for such an environment modeling.

The environment of a vehicle, in the area of autonomous driving, is modeled on the one hand by a feature-based approach with object model assumption, which interprets objects as vectors. On the other hand a grid-based approach is used, which discretizes the environment and all objects in cells and is generally object model free. The environment perception is performed by sensors from which specific information for the representations can be derived with varying degrees of accuracy. 

In feature-based approaches, which typically include object tracking, the intention is usually to track solely dynamic objects, although static objects could also be modeled. An example for this approach is the tracking according to Reuter et al. \cite{Reuter.2014b}, in which the labeled multi-bernoulli (LMB) filter is applied. 
Based on radar measurements for tracking, an object model is assumed for the estimation of pose and movement.

On the contrary, no object model is assumed in the grid-based approach. For this approach there is a static modeling by an occupancy grid map, see \cite{Elfes.1989,Thrun.2006}, or its extension for estimating dynamic objects with the dynamic occupancy grid map (DOGMa), as in \cite{Nuss.2016,Tanzmeister.2014}. So, it is possible to estimate the position in static case and the orientation and movement in dynamic case by a particle filter based on laser measurements.

Both approaches model objects and the environment with different focus. Especially with sparse radar measurements, object hypotheses can be generated by tracking through object model assumptions, whereby continuous tracking is difficult due to a situation. Likewise, the dynamic grid map lacks the expansion of the objects and an accurate tracking of the whole objects, but the form of objects are well presented. 

A method that combines both forms of representation and enables a smooth transition into each other is desirable. Various fusion methods have been proposed for this purpose \cite{Bouzouraa.2010,Gies.2018,Nuss.2014,Nuss.2016,Steyer.2017,Vatavu.2018}. 

The key contribution of this paper is a fusion, which does not make any object assumptions, but uses information from the association of an object hypothesis and grid cells to generate new hypotheses. Therefore, any dynamic objects can be tracked during this fusion. Thus, the fusion algorithm is so modular that the environment modeling is decoupled from it and the logic is purely in the fusion. The evaluation shows that the presented fusion allows a more accurate object estimation and a longer temporal tracking of objects. 

In order to bring the two representations together, a common basis can be found or these can be related, as with the algorithm presented here. A feature vector can serve as a basis. 
In the paper by Steyer et al. \cite{Steyer.2017} object hypotheses are directly created based on the DOGMa. Here, however, no fusion is carried out, but only a hypothesis is created on a grid map. The approach of Bouzouraa \cite{Bouzouraa.2010} exchanges information between models to improve estimates and create a common list of object hypotheses. For this purpose, model information is also exchanged and only the distance measurement is examined. A more comprehensive fusion with additional context information is performed on \cite{Gies.2018, Nuss.2014}. Gies et al. \cite{Gies.2018} proposes an architecture in which, meta object lists are created as a common basis that are fused on the basis of model-based and constrain-based confidence measures. For this purpose, object hypotheses are created from the DOGMa, which, together with context information and object hypotheses, are fused into a high-level fusion. The approach is also structured in such a way that the environment representations are not adapted.

\section{Environment Models}
This section introduces the representations for the two environment models and points out their differences. This is followed by a concept for the fusion of representations. 

\begin{figure}[thpb]
	\centering
	\includegraphics[width=1\columnwidth]{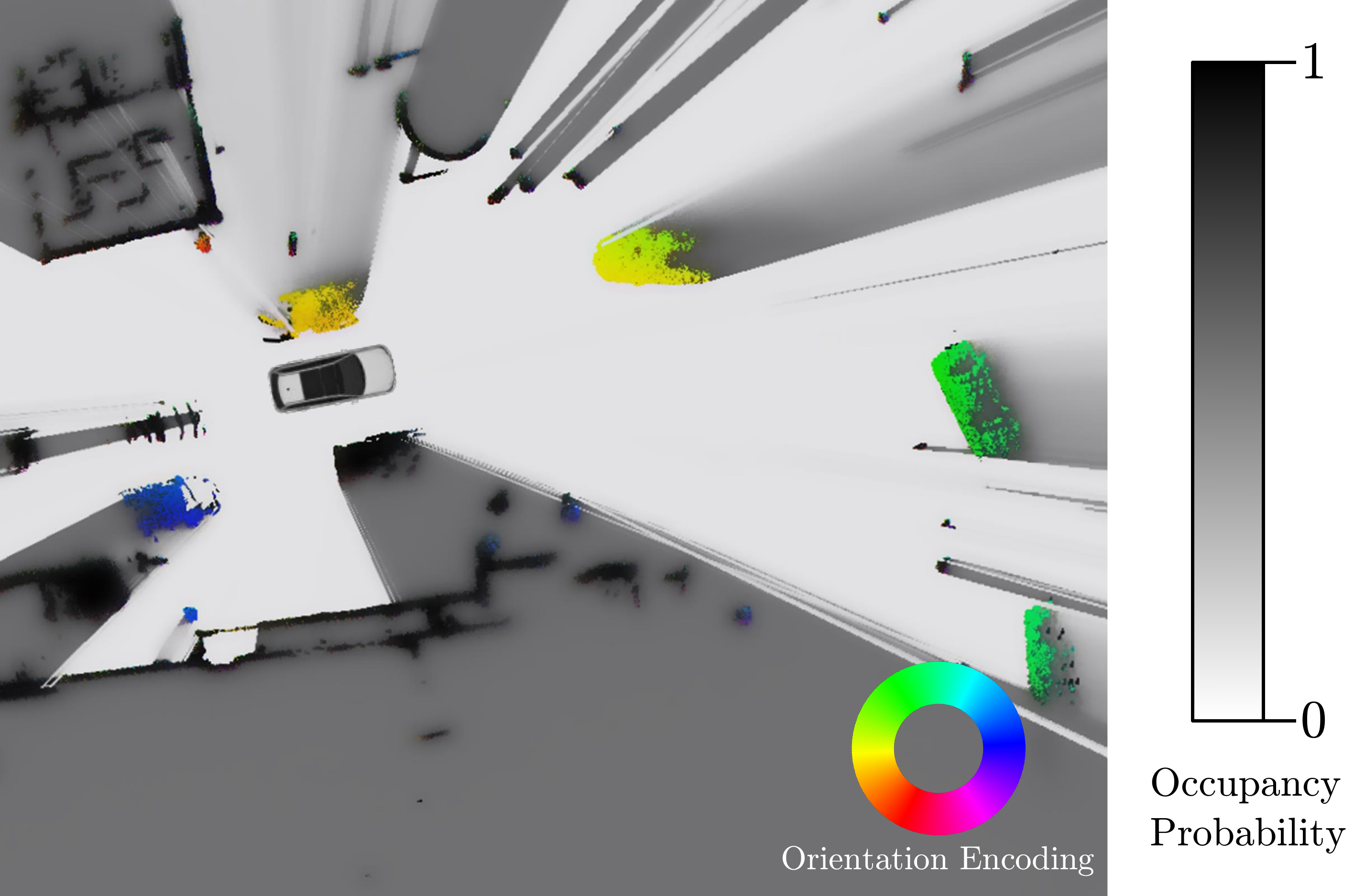}
	\caption{Visualization of a dynamic occupancy grid map (DOGMa)}
	\label{fig:dogma}
\end{figure}

For the grid-based approach, the implementation of the dynamic occupancy grid map (DOGMa) according to Nuss~et~al.~\cite{Nuss.2016} is used. The environment is divided into \SI{15}{\centi\metre} square cells. By this division objects are represented as a set of cell-size point objects, whereby in each cell at most only one such point object exists. Thus, this representation is without object model. As sensor for the DOGMa in this paper an ibeo LUX four layer laser scanner is used with a horizontal angle of view (AOV) of \SI{85}{\degree} with four layers, further \SI{25}{\degree} with two layers, a horizontal angular resolution of \SI{0.25}{\degree}, vertical AOV of \SI{3.2}{\degree} and a range up to \SI{200}{\metre}. Measurements generated by this laser scanner are inserted into a grid map by an inverse sensor model \cite{Thrun.2006}. This measurement grid map serves as an input for the DOGMa, in which the 2D velocity for each cell is estimated by a particle filter in the background. \cref{fig:dogma} shows the DOGMa, in which the  occupancy probability for each cell is represented by the intensity of the gray scale value. The colored cells illustrate the direction of movement of dynamic point objects, correspondingly coded with the color circle at the bottom right.

Based on the subdivision into cells, the DOGMa does not require an explicit object model assumption, but the whole environment.
In general the effort to calculate the particle filter is high and therefore a simple motion model, the constant velocity (CV) model \cite{Schubert.2008}, was chosen to keep the state space for the particle filter small. Accordingly, the particle state space comprises two dimensions each for position and velocity. The sum of the weights of the particles of a cell represents the probability of their occupancy. The state vector for a cell consists of the  occupancy probability and the two-dimensional position as well as velocity.

In the feature-based approach, objects of an environment are modeled as a vector, i.e. a point, in the state space. A representative for this approach is the LMB tracking according to Reuter et al. \cite{Reuter.2014b}, which is used in this paper. To identify objects from the data of a measurement, models are used for object assumption. In this work, the tracking is based on sensor data of a long-range radar (LRR) and an ibeo LUX four layer laser scanner, the same as for DOGMa. As object model, a box for a car is assumed. Specifically, this LMB tracking uses a point object tracking, i.e. each object generates at most one measurement at a time step (see \cite{Granstrom.2017}).

The LRR provides a maximum of one measurement for just one object, which can then be incorporated directly into the LMB tracking. Due to its use as a long-range radar, it has a horizontal AOV of \SI{30}{\degree}, vertical AOV of \SI{5}{\degree} and a range up to \SI{250}{\metre}. This means that objects in the vicinity can only be detected in a small area. Thus, the LRR is designed for the use case of an adaptive cruise control (ACC) and rather for longitudinal tracking. In contrast, the ibeo laser scanner has an AOV of \SI{85}{\degree} with four layers and covers a larger area (see specification above at DOGMa). A laser scanner generates more than only one measurement for an object. In order to get only one point for an object for tracking, we preprocess the laser data with a box fitting. This algorithm is not very robust, so that there are problems with the calculation of the box hypotheses and thus also influences the quality of the object hypotheses. It shall be mentioned again that this is a concept and therefore the algorithms are exchangeable. Only any algorithm for the environment modeling is needed.
Initially, a hypothesis is created based on the radar measurement and updated by both LRR and ibeo laser box hypotheses. Overall, only dynamic objects are tracked here, since this has a direct influence on the vehicle's space of movement.

Due to the feature-based approach and the accompanying object representation as a point, there is scarce data in the environment model. For this reason, only a small amount of object hypotheses needs to be calculated, as opposed to the DOGMa approach where each cell that can be an object is calculated by a set of particles. Therefore, tracking uses a more complex motion model, the CTRV model \cite{Schubert.2008}. 

\begin{figure}[thpb]
	\centering
	\includegraphics[width=\columnwidth]{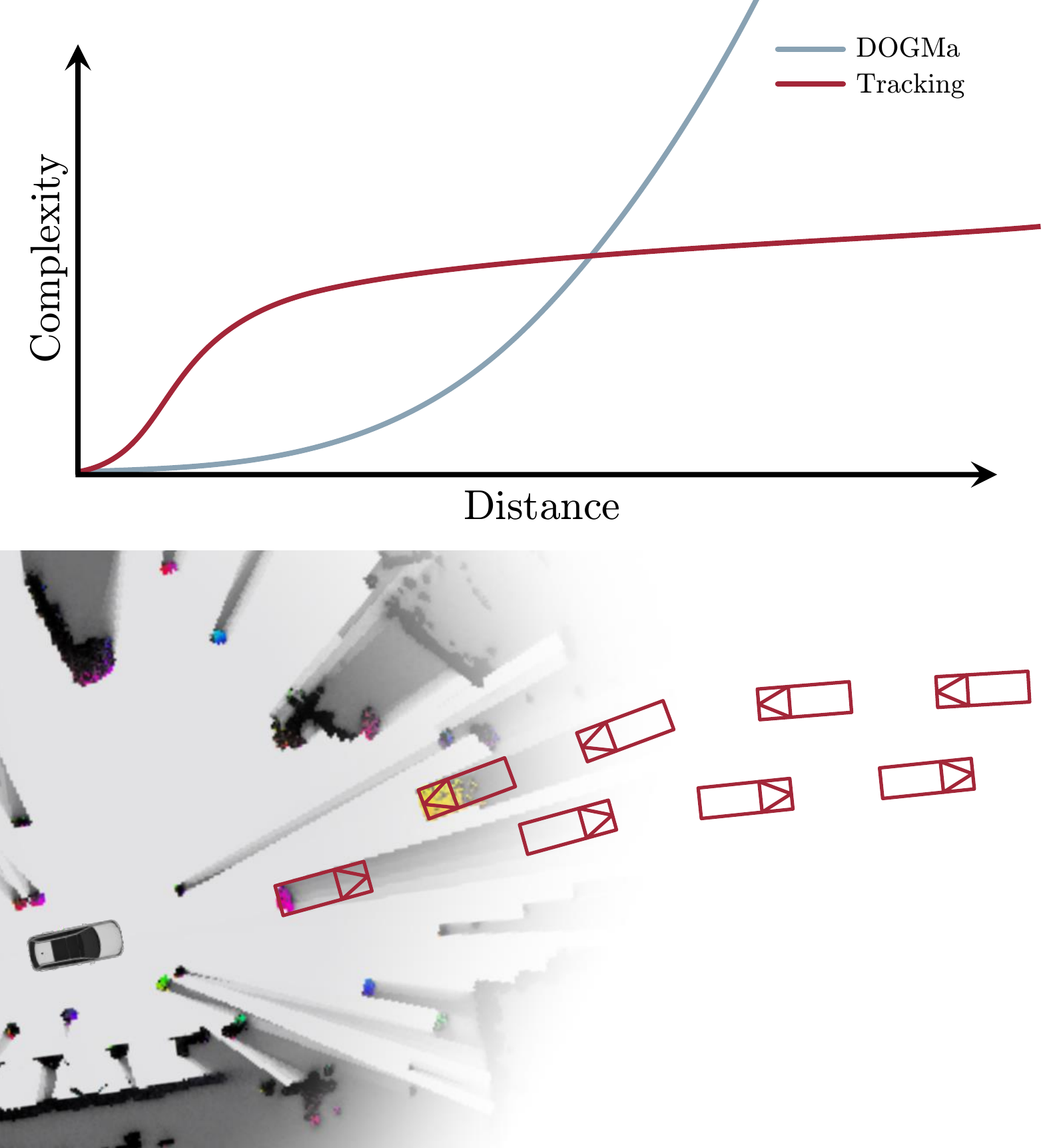}
	\caption{Common illustration of the grid-based (left) and feature-based (right) approach. The graph above visualizes the tendential complexity of these approaches as a function of distance.}
	\label{fig:complexity}
\end{figure}

In summary, the two models in \cref{fig:complexity} are illustrated with reference to their arbitrary complexity over the distance, starting from the ego vehicle. Particularly in the immediate vicinity there are objects with versatile and amorphous forms, such as bushes, artworks and pedestrians with prams, scooters or wheelchairs, which can be captured in detail by high-resolution laser scanners. Due to the high level of detail in the vicinity, many measurements are also available. As the distance increases, the sensor coverage area and thus the number of measurements as well as the perceived details decrease. Radars in particular can still detect objects far in the distance, but provide comparatively fewer measurements than laser scanners.

\cref{fig:complexity} shows that the complexity of the grid-based approach increases approximately squarish with distance. With the feature-based approach, the complexity increases strongly in the near environment and then flattens out with increasing distance, since with the feature-based approach a large number of object models have to be associated in the detailed measurement. In the distance, only a small amount of data is added, so that only a small amount of more object models need to be matched.

The grid-based approach represents the environment as a square grid map, which also explains the quadratic complexity. This has advantages especially in the vicinity. When creating the grid map, no object models are assumed, just the measurements are summarized in discrete areas. Since all objects are divided into cells, any form can be mapped as accurately as possible, depending on the cell size. This increases the calculation effort even if only a few measurements are available, because with this approach the complete environment is modeled.

In contrast, for an object model-based approach, such as the feature-based approaches, it is challenging to assign detailed resolved forms and unstructured data to the given object models. Among other things, a lot of preprocessing effort is required to fit,  e.g. box models, into large amount of data in the vicinity. On the contrary, in distant areas mainly objects such as vehicles are interesting as object models and the data is sparser, a matching of the data to object models is less complex. Areas far away are covered in particular by LRRs that generate only a small amount of measurement data.

In summary, the dynamic grid approach is advantageous for the near area and the object based approach for the far area. It is desirable to have an approach that combines the two methods in order to use the benefits of both.  For the smooth transition of these areas, including the urban area, both environment models should match their information. This is exactly where this work begins and describes a concept for a fusion of the two models.


\section{Algorithm of Fusion}

\subsection{Concept of Fusion}
\begin{figure*}[t]
	\centering
	\includegraphics[width=\textwidth]{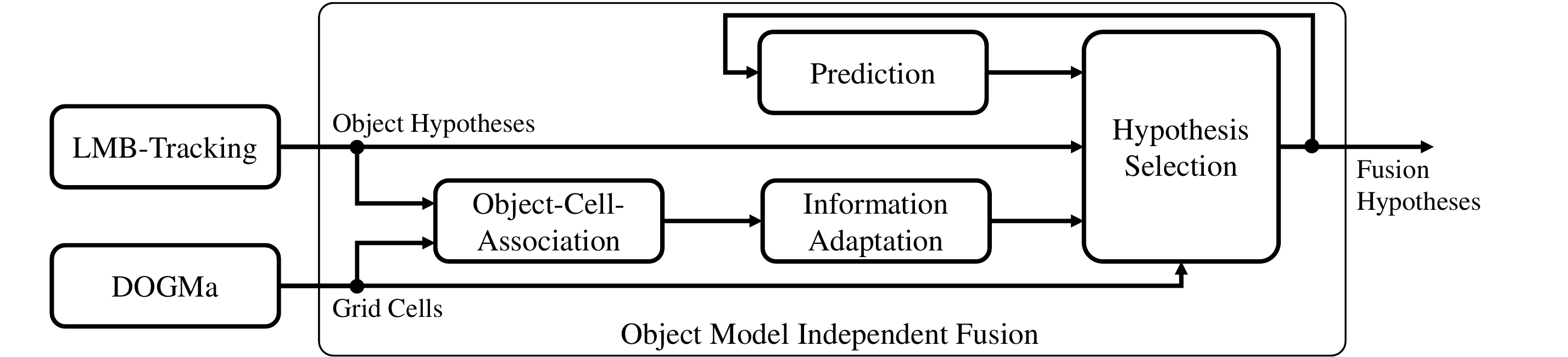}
	\caption{Schematic sequence of the fusion algorithm with the several modules.}
	\label{fig:fusion_model}
\end{figure*}

The algorithm presented in this paper proposes how the information from the DOGMa and LMB tracking techniques can be combined. At the beginning, an overview of the developed algorithm is given before the implementation is discussed in more detail. For clarification it should be mentioned once again that two existing methods are used here and the focus of the presented algorithm is on the possibility of
fusion of information from both techniques. The representations of these two methods therefore serve as input for the developed concept. \cref{fig:fusion_model} schematically illustrates this concept for the fusion algorithm. The scheme shows that the modules of the fusion have no effect on the representations. This fusion is described as an object model independent fusion, since in fusion and especially in information adaptation the individual cells of the DOGMa are used. Thus, the fusion can be applied to different object types.

Information about cells is known from the DOGMa and from tracking there are point object hypotheses, which also describe an extension of the objects. Thus, for a fusion an association between grid cells and object hypotheses is necessary. It should be noted that different information about an object can exist through the environment representations, leading to several hypotheses. For the final result the best fitting hypothesis is selected from the hypothesis of object tracking, fused hypothesis and the predicted hypothesis.
The individual modules are explained in the following.

\subsection{Association Between Grid Cells and Object Hypotheses}

\begin{figure}[t]
	\centering
	\subfloat[Initial object hypothesis of\newline cross traffic] {
		\label{fig:grid_cell_association_example_wrong_ori}
		\includegraphics[width=0.47\columnwidth]{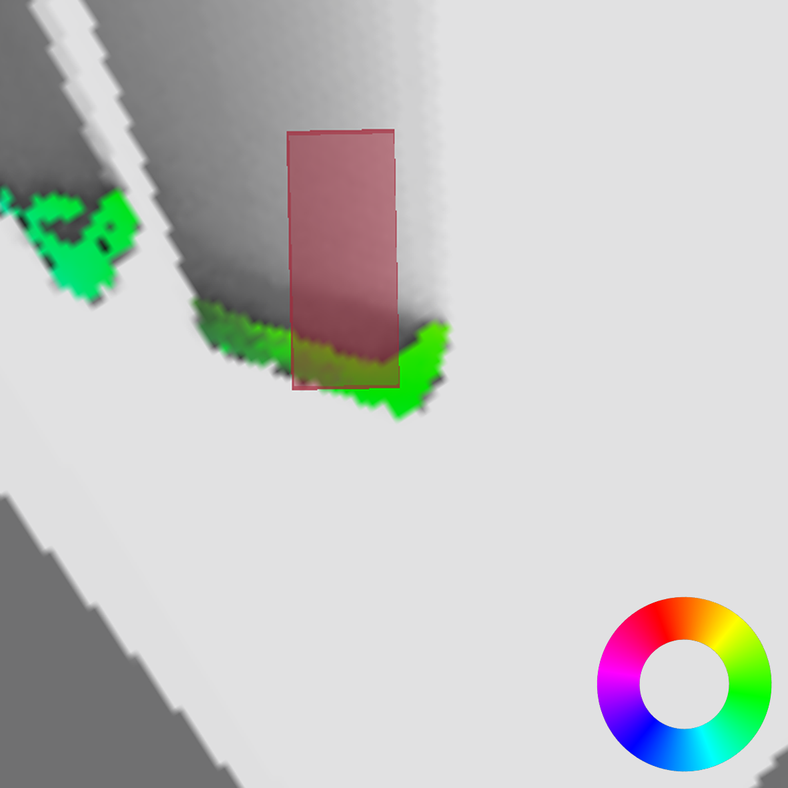}
	}
	\subfloat[Object hypotheses beside object with fusion result (green) ] {
		\label{fig:grid_cell_association_example_beside}
		\includegraphics[width=0.47\columnwidth]{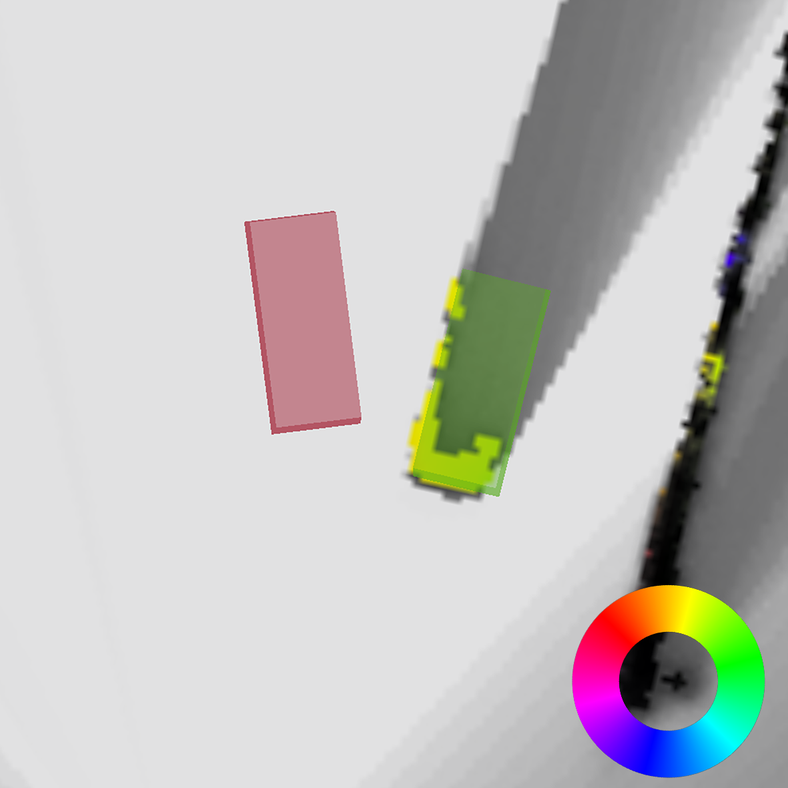}
	}
	\caption{Examples of use cases for the association between object hypotheses (red) and grid cells.}
	\label{fig:grid_cell_association_example}
\end{figure}

At the beginning of the algorithm, the grid cells must be assigned to the tracking object hypothesis. For this purpose, it is possible to generate object hypotheses from the DOGMa, which are then compared with the tracking result, as is done by clustering at \cite{Steyer.2017,Gies.2018}. However, this produces incorrect measurements (e.g. false-positives) which have to be corrected in further steps. Therefore, the corresponding cells of the grid map are identified here on the basis of the object hypotheses for simplification. This is due to the fact that information is available from both techniques for a fusion and the focus is on this. In this way, the cells are examined under the object hypothesis and in the immediate vicinity for dynamic cells, because only the dynamic cells are relevant. In addition to the found cells, all those dynamic cells are added, which are connected to these cells by a 8\hbox{-}connected neighborhood. Dynamic cells are defined as those which have a occupancy probability greater than 0.7 and whose velocity is clearly in one direction, with a design parameter. For the expression of the velocity the distance by the mahalanobis distance between the velocity vector of a cell and the static, i.e. the zero velocity, is used \cite{Nuss.2016}. Two examples of the association is shown in \cref{fig:grid_cell_association_example}. Thereby parts of the DOGMa are pictured and the object hypothesis of tracking as a red box. \Cref{fig:grid_cell_association_example_wrong_ori} illustrates a hypothesis that was initially incorrectly oriented. In the association, the green cells below are linked to this hypothesis by the neighborhood relationship. 

If no cell is found under the object hypothesis, the area larger by a factor of 1.5 of object length is inspected. This is for the case that the rotation or position of the hypothesis does not match the representation of the DOGMa, then an assignment of dynamic cells to the object hypothesis of the tracking can take place nevertheless. This case is shown in the \cref{fig:grid_cell_association_example_beside}. The object hypothesis (red) lies next to the dynamic cells. A green box for the fusion hypothesis is shown. This result is based on the adaptation of the associated cells. The orientation of the hypothesis is fitted to the form in the DOGMa.

\subsection{Realization of Fusion}

\begin{figure}[thpb]
	\centering
	\subfloat[DOGMa with object hypotheses] {
		\label{fig:rotated_hypothesis_dogma}\includegraphics[width=0.47\columnwidth]{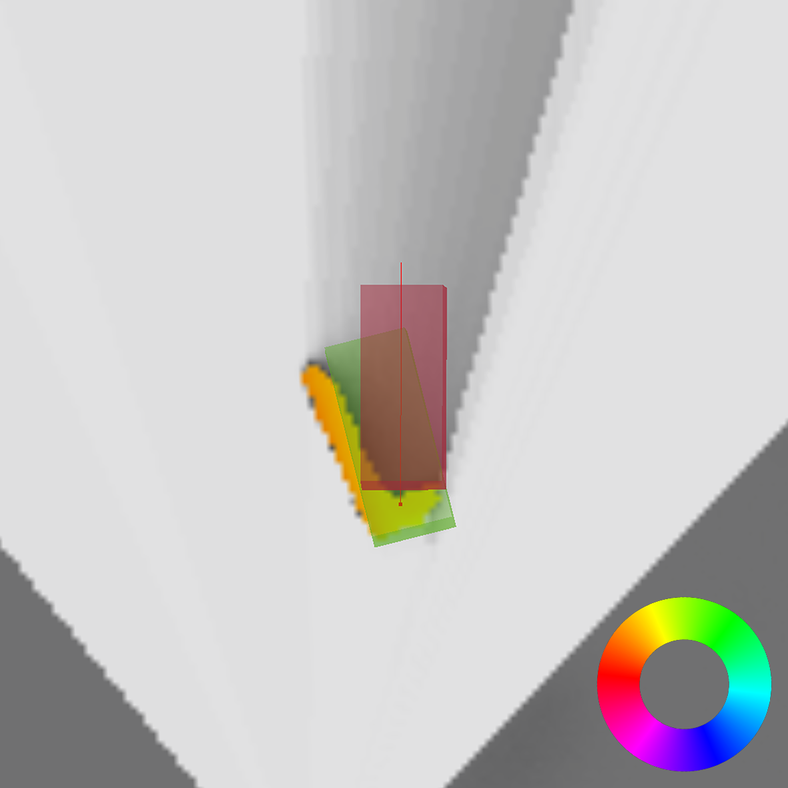}
	}
	\subfloat[Corresponding video frame \newline] {
		\label{fig:rotated_hypothesis_video}
		\includegraphics[width=0.47\columnwidth]{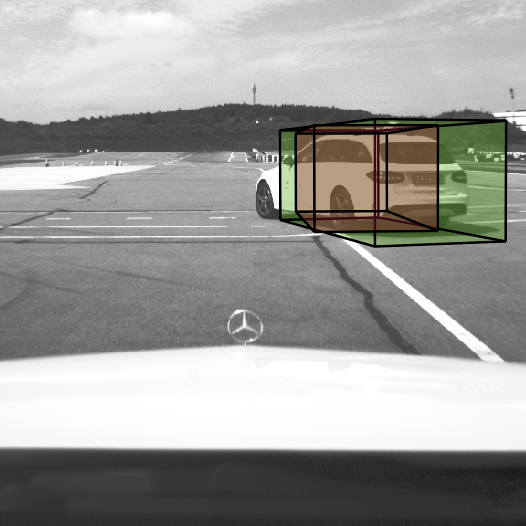}
	}
	\caption{Adaptation of the orientation of the object hypothesis (red) based on the form of the object in the DOGMa. In green the result of the fusion.}
	\label{fig:rotated_hypothesis}
\end{figure}

Due to the representations, the estimated states of the object hypotheses of the tracking may differ from those of the associated cell group. Strictly speaking, the average of velocity and orientation over this cell group is calculated and these may have different values. Especially when initializing object hypotheses of tracking, for example, cross-traffic is set up misaligned due to the point object tracking, the assumption that the car hypotheses are aligned longitudinally and the low coverage of the LRR in the intersection area (see \cref{fig:grid_cell_association_example_wrong_ori}). Whereas based on the object model free approach of DOGMa, the measurements are registered according to the cells and thus reflect the occupancy of the environment independent of object assumptions, whereby the form is discretized in cells. Therefore, the form of the objects is more realistic than represented by the tracking approach. This is presented in \cref{fig:rotated_hypothesis}, with a section of the DOGMa close to the ego vehicle with hypotheses and the corresponding video frame shown.

In both images of \cref{fig:rotated_hypothesis} are illustrated the object hypothesis of tracking in red and those of the fusion in green. This is a sample of sequence \#2 and the object tracking is only using LRR data. In the video frame (\cref{fig:rotated_hypothesis_video}) the hypotheses for illustration are drawn as 3D boxes in the image.  This shows that the hypothesis of the fusion is more realistic. In order to see this in the DOGMa, \cref{fig:rotated_hypothesis_dogma} shows the section at the same time. Additionally there is a red dot with a line in the image. This is an LRR measurement which is used by object tracking. The estimation of the cells goes to the left according to the color circle. Since the vehicle first entered the sensor area of the LRR, a longitudinal hypothesis was first set up by the tracking. For this reason, the information of the associated cell group can be used to adjust the position and orientation of object hypothesis. 

The orientation is determined according to the averaged velocity vector of the cell group and the object hypothesis of tracking is rotated with these. For the position, it is determined which corner point is visible from the vehicle. Starting from this corner point, the hypothesis is shifted from the tracking. 

The estimation of the velocity and thus also the orientation is based on a particle filter, which uses the CV model, resulting in difficulties when driving around curves. Hereby, the orientation  in the direction of the boundary of the curve is estimated. In this case the tracking can better track the orientation due to the used CTRV model. For this reason, the object hypothesis is held as an additional hypothesis. 

The presented approach is based on the existence of an object hypothesis of tracking. This is also necessary for the initialization. However, after the cells have been associated to the hypothesis, the selected hypothesis can be predicted for the next time step and included in the calculation as a possible hypothesis. 

\subsection{Selection of Object Hypothesis}

In the presented approach, two further hypotheses were generated for one object hypothesis. Now a criterion is required for the selection of the more precise hypothesis as to reality. In general, such an evaluation is difficult, as some assumptions have already been made for measurements, movements or objects for the state estimation.  

In the used tracking approach, point objects are tracked based on a single measurement. Therefore, no orientation can be made directly, but rather by movement and model assumptions. In addition, the object is abstracted to a point. In the grid-based approach, as described in the previous section, the form of the objects is discretized by cells and thus a set of cells represents an object. This produces a more realistic representation than a feature-based approach. Consequently, the underlying cells are examined for the selection of the object hypothesis. 

Finally, the number of dynamic cells below an object hypothesis is the decisive criterion. Cells that are regarded as static are excluded because false positive hypotheses are so filtered. 

\section{Evaluation}
\label{sec:evaluation}

The fusion presented in this paper is implemented with C++ and is evaluated by real experimental sequences. The Ulm University own test vehicles are used for this research. A qualitative evaluation is carried out on a sequence of a test area with up to three test vehicles, so that the results can be compared to ground truth data. Each test vehicle is equipped with a differential global positioning system (DGPS) sensor, which ensures maximum accuracy. The ego vehicle is equipped with an ibeo LUX four layer laser scanner and an LRR in the center of the front. The DOGMa uses the laser scanner data. Input for the LMB tracking is either only the LRR data or the ibeo data in combination with a box fitting and the LRR data. 

\begin{figure}[t]
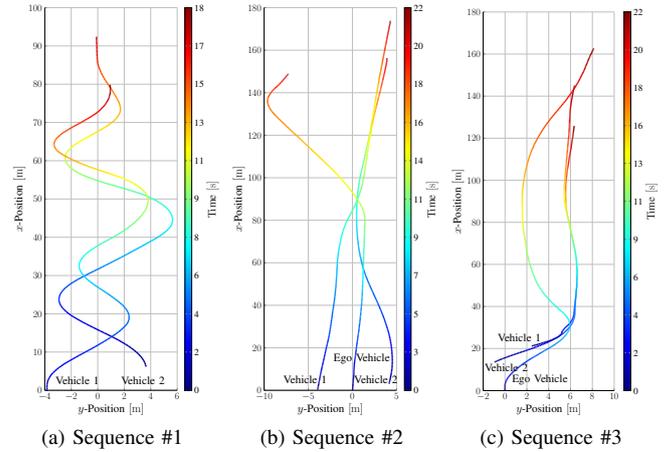

	\centering
	\subfloat[Sequence \#1] {\label{fig:driving_route_seq_1}
	\resizebox{.32\columnwidth}{!}{%
		\input{src/Rec20170824121652_U1800_fusion_LRR+IBEO_DrivingRoute.tex}
	}%
	}
	\subfloat[Sequence \#2] {\label{fig:driving_route_seq_2}
		\resizebox{.32\columnwidth}{!}{%
			\input{src/Rec20170824121051_U1800_fusion_LRR+IBEO_DrivingRoute.tex}
		}%
	}
	\subfloat[Sequence \#3] {\label{fig:driving_route_seq_3}
		\resizebox{.32\columnwidth}{!}{%
			\input{src/Rec20170727125631_U1800_fusion_LRR+IBEO_DrivingRoute.tex}
		}%
	}
	\caption{Driving routes of the reference vehicles 1 and 2 as well as of the ego vehicle over time in color scale. In sequence \#1 is the ego vehicle stationary.}
	\label{fig:driving_routes}
\end{figure}

To illustrate the sequences, the routes of the reference vehicles and the ego vehicle are shown in \cref{fig:driving_routes}. The sequences show complex situations in which the vehicles drive out or into the sensor area. Further, the vehicles obscure each other and the measurements were carried out at a long distance. In the first sequence, no route of the ego vehicle is drawn, because it is stationary. The graphics show the chronological course of the routes via color scale. The positions are given in ego-stationary coordinates. 

The results of the qualitative evaluation are noted in \cref{tbl:evaluation}. The columns are for the respective RMSE for the vehicles V1 and V2 as well as the tracking time of the objects. For each sequence, the values were evaluated for LMB tracking with LRR only and in combination of LRR and ibeo laser scanner. The results for position and orientation are rounded to two and for time to one decimal places. Furthermore, there is a line marked with symbol \%, in which the percentage improvement of the error as well as the percentage extension for the track time are entered.

The developed fusion requires an object hypothesis at the beginning. After the cells have been associated with the hypothesis, the fusion algorithm can hold the hypothesis over a longer period of time. This can be seen in the last columns of \cref{tbl:evaluation}. The duration of tracking is determined by accumulating the times, if the time difference between two adjacent time steps is less than \SI{100}{\milli\second}. This means that the hypotheses must be continuous.

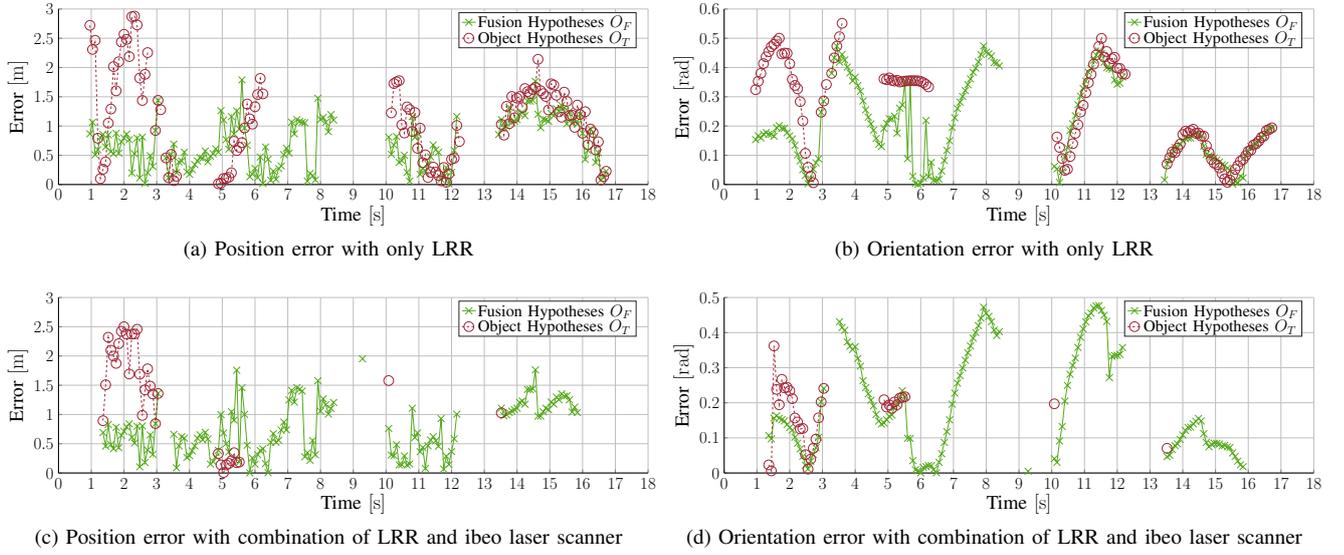
\begin{figure*}[t]
	\centering
	\subfloat[Position error  with only LRR] {
		\label{fig:comparison_seq_1_pos_lrr}
		\resizebox{\columnwidth}{!}{%
%
\definecolor{mycolor1}{rgb}{0.33730,0.66670,0.10980}%
\definecolor{mycolor2}{rgb}{0.63920,0.14900,0.21960}%
\begin{tikzpicture}

\begin{axis}[%
width=10in,
height=3in,
at={(0.833in,0.615in)},
scale only axis,
unbounded coords=jump,
xmin=0,
xmax=18,
xlabel style={font=\bfseries\color{white!15!black}},
xlabel={Time  $[\si{\second}]$},
ymin=0,
ymax=3,
ylabel style={font=\bfseries\color{white!15!black}},
ylabel={Error $[\si{\metre}]$},
axis background/.style={fill=white},
axis x line*=bottom,
axis y line*=left,
xmajorgrids,
ymajorgrids,
legend style={legend cell align=left, align=left, draw=white!15!black, font=\huge},
x tick label style={font=\huge, yshift=-0.5ex},
y tick label style={font=\huge, xshift=-0.5ex},
label style={font=\Huge}
]
\addplot [color=mycolor1, line width=1.2pt, mark=x, mark size=6.0pt, mark options={solid, mycolor1}]
  table[row sep=crcr]{%
0.960013	0.867066273857155\\
1.039988	1.06288171913186\\
1.120014	0.506306884494318\\
1.199997	0.594682144189363\\
1.280023	0.802355808245867\\
1.359998	0.856666830914094\\
1.440024	0.642970989620473\\
1.519998	0.848699592525829\\
1.600025	0.563269988140956\\
1.679999	0.526060988913891\\
1.760025	0.876009944464183\\
1.84	0.536476540398922\\
1.920026	0.731799133683833\\
2.000001	0.794902630272663\\
2.080027	0.861363936263491\\
2.16001	0.765485649985269\\
2.240036	0.197104385961584\\
2.32001	0.528233575910139\\
2.400036	0.793673265797668\\
2.480011	0.113057745314137\\
2.560037	0.811251553181766\\
2.640012	0.0202384196333902\\
2.720038	0.201965573308861\\
2.800013	0.491733469429761\\
2.880039	0.309299380997114\\
2.960014	0.923937583998306\\
3.040041	1.43910923572991\\
nan	nan\\
3.280023	0.455497022711636\\
3.36005	0.115517373599292\\
3.440024	0.514939075942269\\
3.52005	0.690549207304663\\
3.600029	0.205516414053385\\
3.680052	0.386697039922296\\
3.760026	0.401011736425765\\
3.840052	0.551752994925636\\
3.920027	0.216660006025251\\
4.000053	0.17522253689701\\
4.080028	0.25767325295493\\
4.160062	0.341897328212291\\
4.240037	0.401978514180698\\
4.320063	0.461883595481325\\
4.400041	0.4483046129005\\
4.480064	0.572728696375655\\
4.560038	0.388150584921625\\
4.640065	0.483686832756455\\
4.720039	0.52614099115408\\
4.800065	0.610459653522796\\
4.88004	0.553238776445298\\
4.960066	1.26251781008116\\
5.040041	1.10254524245816\\
5.120067	0.767845369412576\\
5.20005	0.604067819004413\\
5.280075	1.16066995074839\\
5.36005	0.86505129642331\\
5.440076	1.25852003781421\\
5.520052	0.948729878143197\\
5.600077	1.78879470143756\\
5.680055	0.966354659801254\\
5.760078	0.67011825656482\\
5.840053	0.132913902668726\\
5.920079	0.178563377762934\\
6.000054	0.277494735733605\\
6.08008	0.121159104196771\\
6.160063	0.470947752464284\\
6.240089	0.0182677339787958\\
6.320063	0.640884507629629\\
6.40009	0.431005216880187\\
6.480064	0.073981981261511\\
6.56009	0.228673167473417\\
6.640066	0.0897356611593167\\
6.720091	0.274381635237809\\
6.800066	0.316914241900989\\
6.880092	0.510715219866185\\
6.960067	0.619580151329281\\
7.040096	0.536455208103739\\
7.120068	1.06533119648951\\
7.200102	0.866292835303042\\
7.280077	1.10949321989126\\
7.360103	1.07736524221405\\
7.440077	1.06208594253918\\
7.520103	1.0564784827555\\
7.600078	0.0592096338448798\\
7.680107	0.189551488211997\\
7.760079	0.152070048100811\\
7.840105	0.0788884857902374\\
7.92008	1.47495861422926\\
8.000107	1.11200185789008\\
8.08008	1.14698244694689\\
8.160115	1.06393461687857\\
8.24009	0.904097553630045\\
8.320116	1.19471376831719\\
8.400092	1.10243091409416\\
nan	nan\\
10.080133	0.809727477160052\\
10.160118	0.515452121082689\\
10.240142	0.675620650462392\\
10.320134	0.820123743298595\\
10.400143	0.377861518645815\\
10.480118	0.409657090007229\\
10.560144	0.500885588809865\\
10.640118	0.208047574548061\\
10.720145	0.0568898920063958\\
10.800119	1.15996680973409\\
10.880148	0.805663307829221\\
10.96012	0.884434161641821\\
11.040146	0.174552740076351\\
11.120121	0.275624979729386\\
11.200154	0.231184403925752\\
11.280129	0.347042693986765\\
11.360157	0.472904780188642\\
11.44013	0.676817359503659\\
11.520156	0.593561818949084\\
11.600131	0.549946049955352\\
11.680157	0.0876482916228269\\
11.760133	0.137610269768743\\
11.840158	0.288067577629178\\
11.920133	0.011824829353543\\
12.000161	0.208427106826903\\
12.080133	0.584803663371929\\
12.160167	1.16315506509852\\
nan	nan\\
13.440182	0.854047540044064\\
13.520157	1.10926414377427\\
13.600183	0.976976028955988\\
13.680158	0.99535725311722\\
13.760184	1.31782956560254\\
13.840159	1.05717775130246\\
13.920184	1.08246406120232\\
14.000159	1.10693597355917\\
14.080186	1.20558074032928\\
14.160168	1.23132701028572\\
14.240197	1.17579951492942\\
14.320169	1.4280796693308\\
14.400195	1.42608929888326\\
14.48017	1.4388534775082\\
14.560196	1.76249459080731\\
14.640171	1.19964643834615\\
14.720197	0.972082964586761\\
14.800171	1.11737678933927\\
14.880197	1.10738803030826\\
14.960172	1.08254720250113\\
15.040198	1.21330938200047\\
15.120173	1.17761993925033\\
15.200206	1.29774675158363\\
15.28018	1.25318715755455\\
15.360206	1.35353459854451\\
15.440181	1.31109478988894\\
15.52021	1.03789601448915\\
15.600182	1.25106353715306\\
15.680208	1.04268541151453\\
15.760183	1.12682586230085\\
15.840209	1.03389573624055\\
15.920184	1.19678116791054\\
16.00021	0.876522461896698\\
16.080184	0.430297206872979\\
16.16022	0.582822179197393\\
16.240195	0.949422709609422\\
16.320221	0.889665706039921\\
16.400197	0.379155649240502\\
16.480222	0.543782979569869\\
16.560197	0.0775726292729928\\
16.640223	0.117221101949411\\
16.720198	0.227237016528377\\
};
\addlegendentry{Fusion Hypotheses $O_F$}

\addplot [color=mycolor2, dashed, line width=1.2pt, mark=o, mark size=6.0pt, mark options={solid, mycolor2}]
  table[row sep=crcr]{%
0.960013	2.72047891520653\\
1.039988	2.30912657525834\\
1.120014	2.46450333075833\\
1.199997	0.791752157571739\\
1.280023	0.101352371515205\\
1.359998	0.261753016448946\\
1.440024	0.387698389550447\\
1.519998	1.04943133978741\\
1.600025	1.2900501836116\\
1.679999	2.01577891242305\\
1.760025	1.60120714771119\\
1.84	2.09914016427424\\
1.920026	2.43536554754381\\
2.000001	2.56825754022064\\
2.080027	2.4805975072072\\
2.16001	2.18689695406636\\
2.240036	2.86923504356123\\
2.32001	2.87875059007153\\
2.400036	2.72718374741826\\
2.480011	1.81994958271911\\
2.560037	1.43686500146961\\
2.640012	1.88320977765704\\
2.720038	2.25504935903191\\
nan	nan\\
2.960014	0.923937583998306\\
3.040041	1.43910923572991\\
3.120017	1.2786798913016\\
nan	nan\\
3.280023	0.455497022711636\\
3.36005	0.115517373599292\\
3.440024	0.514939075942269\\
3.52005	0.0661289671545404\\
3.600029	0.141347186855377\\
nan	nan\\
4.88004	0.0142868458010028\\
4.960066	0.0238920420769446\\
5.040041	0.104223235073924\\
5.120067	0.104444558265016\\
5.20005	0.122894965660052\\
5.280075	0.200421106500151\\
5.36005	0.738448021916831\\
5.440076	0.580851075214198\\
5.520052	0.636961192867204\\
5.600077	0.715308012914974\\
5.680055	0.966354659801254\\
5.760078	1.37632164132216\\
5.840053	1.12215413783002\\
5.920079	1.03207271556766\\
6.000054	1.32954467900218\\
6.08008	1.53813963451934\\
6.160063	1.81424349533822\\
6.240089	1.54939889648544\\
nan	nan\\
10.160118	1.22434629306566\\
10.240142	1.7259649242296\\
10.320134	1.75567937672605\\
10.400143	1.77594162235663\\
10.480118	1.01934459014068\\
10.560144	0.879902497086974\\
10.640118	1.32663930480936\\
10.720145	1.28212122371997\\
10.800119	0.906181519171\\
10.880148	1.23101733462026\\
10.96012	0.618662835111081\\
11.040146	0.96380179620563\\
11.120121	0.373470220472441\\
11.200154	0.507166078178727\\
11.280129	0.119715393154706\\
11.360157	0.209730010232406\\
11.44013	0.298589396954142\\
11.520156	0.210625270454074\\
11.600131	0.155297473752455\\
11.680157	0.0623706378497531\\
11.760133	0.290913910631502\\
11.840158	0.0383594845291526\\
11.920133	0.183352126673398\\
12.000161	0.431263590333728\\
12.080133	0.449366887726569\\
12.160167	1.00917897945317\\
12.240142	0.735384357928424\\
nan	nan\\
13.520157	1.0290109113476\\
13.600183	0.844446736877302\\
13.680158	1.33663870616819\\
13.760184	1.03987617849644\\
13.840159	1.50545339408075\\
13.920184	1.13979103020971\\
14.000159	1.47602477062341\\
14.080186	1.5051758102596\\
14.160168	1.31606693289486\\
14.240197	1.64065451483491\\
14.320169	1.5876992467487\\
14.400195	1.5531068076225\\
14.48017	1.62543353843508\\
14.560196	1.66383804416034\\
14.640171	2.1404719769246\\
14.720197	1.60302888383441\\
14.800171	1.55334231515033\\
14.880197	1.50422667663727\\
14.960172	1.2759516569853\\
15.040198	1.7221966994768\\
15.120173	1.70124019975115\\
15.200206	1.52546957500503\\
15.28018	1.24222453528608\\
15.360206	1.15055561812507\\
15.440181	1.57878974547106\\
15.52021	1.42039908744874\\
15.600182	1.17835328859846\\
15.680208	1.18634595592118\\
15.760183	0.977606576734686\\
15.840209	1.35811977818659\\
15.920184	1.19678116791054\\
16.00021	0.876522461896698\\
16.080184	1.24481099790289\\
16.16022	0.6787034574605\\
16.240195	0.949422709609422\\
16.320221	0.889665706039921\\
16.400197	0.591254226476508\\
16.480222	0.730810569286632\\
16.560197	0.0775726292729928\\
16.640223	0.136564639162529\\
16.720198	0.227237016528377\\
};
\addlegendentry{Object Hypotheses $O_T$}

\end{axis}
\end{tikzpicture}%
		}%
	}
	\subfloat[Orientation error with only LRR] {\label{fig:comparison_seq_1_ori_lrr}
		\resizebox{\columnwidth}{!}{%
%
\definecolor{mycolor1}{rgb}{0.33730,0.66670,0.10980}%
\definecolor{mycolor2}{rgb}{0.63920,0.14900,0.21960}%
\begin{tikzpicture}

\begin{axis}[%
width=10in,
height=3in,
at={(0.833in,0.615in)},
scale only axis,
unbounded coords=jump,
xmin=0,
xmax=18,
xlabel style={font=\bfseries\color{white!15!black}},
xlabel={Time $[\si{\second}]$},
ymin=0,
ymax=0.6,
ylabel style={font=\bfseries\color{white!15!black}},
ylabel={Error $[\si{\radian}]$},
axis background/.style={fill=white},
axis x line*=bottom,
axis y line*=left,
xmajorgrids,
ymajorgrids,
legend style={legend cell align=left, align=left, draw=white!15!black, font=\huge},
x tick label style={font=\huge, yshift=-0.5ex},
y tick label style={font=\huge, xshift=-0.5ex},
label style={font=\Huge}
]
\addplot [color=mycolor1, line width=1.2pt, mark=x, mark size=6.0pt, mark options={solid, mycolor1}]
  table[row sep=crcr]{%
0.960013	0.153350434791719\\
1.039988	0.166337986580732\\
1.120014	0.159173506772337\\
1.199997	0.17216369790702\\
1.280023	0.177485916854046\\
1.359998	0.172121326089986\\
1.440024	0.172045131906977\\
1.519998	0.165248227562014\\
1.600025	0.194203948045871\\
1.679999	0.201371471310157\\
1.760025	0.191740118432449\\
1.84	0.192486442560092\\
1.920026	0.17708017430455\\
2.000001	0.166607996736516\\
2.080027	0.155697292922909\\
2.16001	0.134090854843535\\
2.240036	0.101463325334358\\
2.32001	0.0809474327743978\\
2.400036	0.0509034244790918\\
2.480011	0.0309980694327194\\
2.560037	0.00274824161873966\\
2.640012	0.0195139805873836\\
2.720038	0.0517618308696655\\
2.800013	0.0742369354819878\\
2.880039	0.0881271647079025\\
2.960014	0.247263373964422\\
3.040041	0.285617519396023\\
nan	nan\\
3.280023	0.380016595927373\\
3.36005	0.432999569141018\\
3.440024	0.473129282715938\\
3.52005	0.418415148685355\\
3.600029	0.44488769004252\\
3.680052	0.426080141435946\\
3.760026	0.401642145893835\\
3.840052	0.38157659938728\\
3.920027	0.354469948051914\\
4.000053	0.366964578646176\\
4.080028	0.338150494768378\\
4.160062	0.318819491728778\\
4.240037	0.285789500341494\\
4.320063	0.255373008447286\\
4.400041	0.230188770224867\\
4.480064	0.207177679070362\\
4.560038	0.185766535092945\\
4.640065	0.157455070428238\\
4.720039	0.137912938151821\\
4.800065	0.129940673273616\\
4.88004	0.191826639175044\\
4.960066	0.208663561297605\\
5.040041	0.225294868964319\\
5.120067	0.223725439861241\\
5.20005	0.233777792568368\\
5.280075	0.175556552077644\\
5.36005	0.260969493691437\\
5.440076	0.272441797326366\\
5.520052	0.353346019944389\\
5.600077	0.088336374078569\\
5.680055	0.354917985902581\\
5.760078	0.0282165117499034\\
5.840053	0.00356891995147635\\
5.920079	0.000541781504214356\\
6.000054	0.00379578625000931\\
6.08008	0.0226318981008333\\
6.160063	0.218924455451592\\
6.240089	0.0273483226909157\\
6.320063	0.0751536851668924\\
6.40009	0.0145382004174657\\
6.480064	0.0146991054690346\\
6.56009	0.0179029331426896\\
6.640066	0.0464127671812555\\
6.720091	0.0820780574371547\\
6.800066	0.116929609304652\\
6.880092	0.149550823984391\\
6.960067	0.194954685511863\\
7.040096	0.228264279902984\\
7.120068	0.259726209034557\\
7.200102	0.281477468291033\\
7.280077	0.301013629486722\\
7.360103	0.321025667463299\\
7.440077	0.34150254261063\\
7.520103	0.361192353045308\\
7.600078	0.38388800608281\\
7.680107	0.398184468672849\\
7.760079	0.419141329713987\\
7.840105	0.444327690783254\\
7.92008	0.473919935395138\\
8.000107	0.45264542853586\\
8.08008	0.447295254143231\\
8.160115	0.439349481876146\\
8.24009	0.416754853757787\\
8.320116	0.415724836048032\\
8.400092	0.405508620646188\\
nan	nan\\
10.080133	0.0621618770828221\\
10.160118	0.053304006306309\\
10.240142	0.00675623371539658\\
10.320134	0.0449200284266205\\
10.400143	0.09417805067255\\
10.480118	0.125942494665306\\
10.560144	0.184794547519477\\
10.640118	0.207407866033634\\
10.720145	0.235127283819029\\
10.800119	0.272837244760264\\
10.880148	0.30200058341514\\
10.96012	0.328316062042617\\
11.040146	0.353818331312947\\
11.120121	0.383709365264675\\
11.200154	0.396333617352859\\
11.280129	0.424929075093235\\
11.360157	0.442056942845831\\
11.44013	0.456262509793344\\
11.520156	0.448158236111531\\
11.600131	0.41796305172474\\
11.680157	0.399282872604179\\
11.760133	0.396760911579914\\
11.840158	0.394305532817238\\
11.920133	0.357397925401372\\
12.000161	0.342641082593826\\
12.080133	0.346551885611973\\
12.160167	0.36788414064296\\
nan	nan\\
13.440182	0.015832442201404\\
13.520157	0.0690279897930033\\
13.600183	0.0864518396754725\\
13.680158	0.10740200887831\\
13.760184	0.126912489616281\\
13.840159	0.124306015366628\\
13.920184	0.13758492032964\\
14.000159	0.149861785500441\\
14.080186	0.157022442066527\\
14.160168	0.159056147136631\\
14.240197	0.158524852084167\\
14.320169	0.170196409340097\\
14.400195	0.174322742322854\\
14.48017	0.168808417771707\\
14.560196	0.166426951849303\\
14.640171	0.131153901347258\\
14.720197	0.10311599432357\\
14.800171	0.0982578033048291\\
14.880197	0.0946586844349895\\
14.960172	0.0909515108239596\\
15.040198	0.084346137314288\\
15.120173	0.0771745658982468\\
15.200206	0.0673478270460279\\
15.28018	0.0651776960600614\\
15.360206	0.0557514171936404\\
15.440181	0.0491726297676642\\
15.52021	0.00168755756410377\\
15.600182	0.0098177719372643\\
15.680208	0.00604002091403544\\
15.760183	0.0282265859704509\\
15.840209	0.0226215056387922\\
15.920184	0.0987465325149302\\
16.00021	0.113653559284607\\
16.080184	0.129880099649544\\
16.16022	0.13659413380725\\
16.240195	0.139440188389212\\
16.320221	0.148347063314958\\
16.400197	0.16393746127941\\
16.480222	0.168626175306802\\
16.560197	0.184958795573692\\
16.640223	0.1924395894127\\
16.720198	0.193835473805841\\
};
\addlegendentry{Fusion Hypotheses $O_F$}

\addplot [color=mycolor2, dashed, line width=1.2pt, mark=o, mark size=6.0pt, mark options={solid, mycolor2}]
  table[row sep=crcr]{%
0.960013	0.324245702780928\\
1.039988	0.352859988405665\\
1.120014	0.380217488278337\\
1.199997	0.413274528607001\\
1.280023	0.436549130432425\\
1.359998	0.454106886959796\\
1.440024	0.47127343928042\\
1.519998	0.479096342238599\\
1.600025	0.490730193171181\\
1.679999	0.500145761875622\\
1.760025	0.44634374913943\\
1.84	0.449361301316866\\
1.920026	0.448767094361081\\
2.000001	0.413048315264177\\
2.080027	0.359300790669995\\
2.16001	0.338759631918702\\
2.240036	0.32746098543539\\
2.32001	0.283346668930307\\
2.400036	0.216993407167941\\
2.480011	0.106625206419068\\
2.560037	0.0589004869440366\\
2.640012	0.0279819593353707\\
2.720038	0.00621941415056071\\
nan	nan\\
2.960014	0.247263373964422\\
3.040041	0.285617519396023\\
3.120017	0.341872472170174\\
nan	nan\\
3.280023	0.380016595927373\\
3.36005	0.432999569141018\\
3.440024	0.473129282715938\\
3.52005	0.505336265870902\\
3.600029	0.551206734673698\\
nan	nan\\
4.88004	0.361244033895328\\
4.960066	0.358281030765653\\
5.040041	0.364498273935418\\
5.120067	0.355990413530247\\
5.20005	0.353724251416704\\
5.280075	0.35639390596436\\
5.36005	0.350411373922526\\
5.440076	0.352175275597844\\
5.520052	0.353490890657202\\
5.600077	0.354337921475931\\
5.680055	0.354917985902581\\
5.760078	0.354957347573009\\
5.840053	0.354782892610302\\
5.920079	0.354414711132666\\
6.000054	0.352403928948046\\
6.08008	0.348507601853203\\
6.160063	0.342312401050744\\
6.240089	0.333686452177677\\
nan	nan\\
10.160118	0.162612332281057\\
10.240142	0.126813069914562\\
10.320134	0.0888811267955205\\
10.400143	0.0484318676769497\\
10.480118	0.0525099640978262\\
10.560144	0.0957903422416457\\
10.640118	0.139528881699686\\
10.720145	0.174851582928341\\
10.800119	0.204160028480359\\
10.880148	0.247666109486218\\
10.96012	0.270802463316766\\
11.040146	0.313576936975963\\
11.120121	0.34819278149374\\
11.200154	0.376931300069408\\
11.280129	0.411650820757835\\
11.360157	0.443274091189146\\
11.44013	0.472401393069575\\
11.520156	0.499490733829017\\
11.600131	0.435522794814538\\
11.680157	0.454790254873771\\
11.760133	0.414099821781488\\
11.840158	0.43026246381886\\
11.920133	0.434856221016975\\
12.000161	0.398088245526534\\
12.080133	0.396820845270902\\
12.160167	0.378160803942141\\
12.240142	0.377268835386214\\
nan	nan\\
13.520157	0.0702871050986928\\
13.600183	0.0916973077865073\\
13.680158	0.111403018672636\\
13.760184	0.108745621348103\\
13.840159	0.124486482788065\\
13.920184	0.138260654318476\\
14.000159	0.171280670808937\\
14.080186	0.182182117121709\\
14.160168	0.177291833138618\\
14.240197	0.184087351978786\\
14.320169	0.1890178313514\\
14.400195	0.175093679770661\\
14.48017	0.176128685291232\\
14.560196	0.165634384630771\\
14.640171	0.165394567078987\\
14.720197	0.132460014373574\\
14.800171	0.108792891593893\\
14.880197	0.0858020467899543\\
14.960172	0.0729991704702648\\
15.040198	0.0685691774919412\\
15.120173	0.0565616310285861\\
15.200206	0.0374298320349844\\
15.28018	0.019535396130042\\
15.360206	0.00778915597081564\\
15.440181	0.015560518989993\\
15.52021	0.0319318669478008\\
15.600182	0.0516415737680802\\
15.680208	0.0644091795285995\\
15.760183	0.0800103845677178\\
15.840209	0.0854227824544509\\
15.920184	0.0987465325149302\\
16.00021	0.113653559284607\\
16.080184	0.117197571546143\\
16.16022	0.135105492173718\\
16.240195	0.139440188389212\\
16.320221	0.148347063314958\\
16.400197	0.160659983260491\\
16.480222	0.165737825896688\\
16.560197	0.184958795573692\\
16.640223	0.188533062329384\\
16.720198	0.193835473805841\\
};
\addlegendentry{Object Hypotheses $O_T$}

\end{axis}
\end{tikzpicture}%
		}%
	}
	\\
	\subfloat[Position error with combination of LRR and ibeo laser scanner] {\label{fig:comparison_seq_1_pos_lrr_ibeo}
		\resizebox{\columnwidth}{!}{%
%
\definecolor{mycolor1}{rgb}{0.33730,0.66670,0.10980}%
\definecolor{mycolor2}{rgb}{0.63920,0.14900,0.21960}%
\begin{tikzpicture}

\begin{axis}[%
width=10in,
height=3in,
at={(0.833in,0.615in)},
scale only axis,
unbounded coords=jump,
xmin=0,
xmax=18,
xlabel style={font=\color{white!15!black}},
xlabel={Time $[\si{\second}]$},
ymin=0,
ymax=3,
ylabel style={font=\color{white!15!black}},
ylabel={Error $[\si{\metre}]$},
axis background/.style={fill=white},
axis x line*=bottom,
axis y line*=left,
xmajorgrids,
ymajorgrids,
legend style={legend cell align=left, align=left, draw=white!15!black, font=\huge},
x tick label style={font=\huge, yshift=-0.5ex},
y tick label style={font=\huge, xshift=-0.5ex},
label style={font=\Huge}
]
\addplot [color=mycolor1, line width=1.2pt, mark=x, mark size=6.0pt, mark options={solid, mycolor1}]
  table[row sep=crcr]{%
1.359998	0.690916610648001\\
1.440024	0.45951175663091\\
1.519998	0.829130417579343\\
1.600025	0.485618893441295\\
1.679999	0.414858216953753\\
1.760025	0.781513295389153\\
1.84	0.432851494746568\\
1.920026	0.643427307393743\\
2.000001	0.713612784963505\\
2.080027	0.789247123062513\\
2.16001	0.841197684528669\\
2.240036	0.611720864393629\\
2.32001	0.511889152082084\\
2.400036	0.799124913214435\\
2.480011	0.105641217959239\\
2.560037	0.805442687515001\\
2.640012	0.180395937659021\\
2.720038	0.391548577100433\\
2.800013	0.650886312366605\\
2.880039	0.318940683619243\\
2.960014	0.843946569058748\\
3.040041	1.35911822079035\\
nan	nan\\
3.52005	0.660995465450732\\
3.600029	0.0898917242886128\\
3.680052	0.462783883920885\\
3.760026	0.621867228412306\\
3.840052	0.595168574900647\\
3.920027	0.266749895448513\\
4.000053	0.296204466210675\\
4.080028	0.378655251143627\\
4.160062	0.462878716733414\\
4.240037	0.593851182844375\\
4.320063	0.653756663335372\\
4.400041	0.569286506057587\\
4.480064	0.693711091728112\\
4.560038	0.580023770595211\\
4.640065	0.151494446422042\\
4.720039	0.194548909748029\\
4.800065	0.277399654770704\\
4.88004	0.387126472023184\\
4.960066	0.998203596214346\\
5.040041	0.683686240686491\\
5.120067	0.500951483346178\\
5.20005	0.35892070408007\\
5.280075	1.04732693800726\\
5.36005	0.903735130530869\\
5.440076	1.75720170354146\\
5.520052	0.186634023633651\\
5.600077	1.45859773970862\\
5.680055	1.00269000053575\\
5.760078	0.474663163635981\\
5.840053	0.00230627686611484\\
5.920079	0.257207088058493\\
6.000054	0.150612407328843\\
6.08008	0.313082656797519\\
6.160063	0.382634876220649\\
6.240089	0.412884408516604\\
6.320063	0.160296731790069\\
6.40009	0.00490608604944498\\
6.480064	0.524303511852473\\
6.56009	0.673596541981397\\
6.640066	0.524577757901461\\
6.720091	0.559184745571855\\
6.800066	0.733603564047087\\
6.880092	0.87001898876499\\
6.960067	0.737262290929472\\
7.040096	1.24942380819962\\
7.120068	1.41123567043226\\
7.200102	1.21189835494021\\
7.280077	1.46027148275282\\
7.360103	1.43180080451948\\
7.440077	1.40138885583205\\
7.520103	0.276903635494953\\
7.600078	0.346695438037109\\
7.680107	0.217187797173999\\
7.760079	0.553725607885255\\
7.840105	0.312936839959043\\
7.92008	1.57826803151509\\
8.000107	1.05752167901818\\
8.08008	1.26796442160339\\
8.160115	1.16724396429918\\
8.24009	1.00740723232278\\
8.320116	1.10302351783903\\
8.400092	1.20574061404135\\
nan	nan\\
9.280129	1.95299831267237\\
nan	nan\\
10.080133	0.760630888337872\\
10.160118	0.307249586382824\\
10.240142	0.297349478759656\\
10.320134	0.484149406254772\\
10.400143	0.138801928393775\\
10.480118	0.141076996322269\\
10.560144	0.292061297763581\\
10.640118	0.138752184273201\\
10.720145	0.153819106735696\\
10.800119	1.10608940185024\\
10.880148	0.606220100478865\\
10.96012	0.6865440415731\\
11.040146	0.362992436880901\\
11.120121	0.435407072051788\\
11.200154	0.0784845014983091\\
11.280129	0.465918764928986\\
11.360157	0.563223763947306\\
11.44013	0.620996987136962\\
11.520156	0.551082432843558\\
11.600131	0.45740262233835\\
11.680157	0.929487285020116\\
11.760133	0.0667188902293958\\
11.840158	0.288067577629178\\
11.920133	0.145966294824447\\
12.000161	0.366218261106553\\
12.080133	0.584803663371929\\
12.160167	1.00536421406589\\
nan	nan\\
13.520157	1.10926414377428\\
13.600183	0.976976028955988\\
13.680158	0.995357253117248\\
13.760184	1.03426693793135\\
13.840159	1.05717743043863\\
13.920184	1.08246406120227\\
14.000159	1.10693597355915\\
14.080186	1.20558074032928\\
14.160168	1.23132701028572\\
14.240197	1.17579951492941\\
14.320169	1.42807966933078\\
14.400195	1.42608929888326\\
14.48017	1.4388534775081\\
14.560196	1.76249459080719\\
14.640171	0.970964852057037\\
14.720197	0.972082964586548\\
14.800171	1.04648535849779\\
14.880197	1.1073880303084\\
14.960172	1.08254720250127\\
15.040198	1.21330938200069\\
15.120173	1.17761993925059\\
15.200206	1.29774675158379\\
15.28018	1.25318715755458\\
15.360206	1.35353459854454\\
15.440181	1.31109478988897\\
15.52021	1.32146048380162\\
15.600182	1.25106353715294\\
15.680208	1.0426854115142\\
15.760183	1.12682586230058\\
15.840209	1.03389573624044\\
};
\addlegendentry{Fusion Hypotheses $O_F$}

\addplot [color=mycolor2, dashed, line width=1.2pt, mark=o, mark size=6.0pt, mark options={solid, mycolor2}]
  table[row sep=crcr]{%
1.359998	0.893719502653871\\
1.440024	1.50856521161817\\
1.519998	2.32005930604137\\
1.600025	2.09780096029382\\
1.679999	1.99853356273918\\
1.760025	1.87349488801115\\
1.84	2.21034231149306\\
1.920026	2.42264678073246\\
2.000001	2.50028375193151\\
2.080027	2.37302582825891\\
2.16001	1.69211262067818\\
2.240036	2.37445071017305\\
2.32001	2.37657247195782\\
2.400036	2.45664151553948\\
2.480011	1.68806582028972\\
2.560037	0.984260674490237\\
2.640012	1.42096985612485\\
2.720038	1.78065707847731\\
2.800013	1.48458799471118\\
2.880039	1.35026586116247\\
2.960014	0.843946569058748\\
3.040041	1.35911822079035\\
nan	nan\\
4.88004	0.332389838358822\\
4.960066	0.139110349890395\\
5.040041	0.00800237261490366\\
5.120067	0.13749126504436\\
5.20005	0.163303865059284\\
5.280075	0.209929304449425\\
5.36005	0.346915590635916\\
5.440076	0.176504616839551\\
5.520052	0.186634023633651\\
nan	nan\\
10.080133	1.57951707205192\\
nan	nan\\
13.520157	1.0290109113476\\
};
\addlegendentry{Object Hypotheses $O_T$}

\end{axis}
\end{tikzpicture}%
		}%
	}
	\subfloat[Orientation error with combination of LRR and ibeo laser scanner] {\label{fig:comparison_seq_1_ori_lrr_ibeo}
		\resizebox{\columnwidth}{!}{%
%
\definecolor{mycolor1}{rgb}{0.33730,0.66670,0.10980}%
\definecolor{mycolor2}{rgb}{0.63920,0.14900,0.21960}%
\begin{tikzpicture}

\begin{axis}[%
width=10in,
height=3in,
at={(0.833in,0.615in)},
scale only axis,
unbounded coords=jump,
xmin=0,
xmax=18,
xlabel style={font=\bfseries\color{white!15!black}},
xlabel={Time $[\si{\second]}$},
ymin=0,
ymax=0.5,
ylabel style={font=\bfseries\color{white!15!black}},
ylabel={Error $[\si{\radian}]$},
axis background/.style={fill=white},
axis x line*=bottom,
axis y line*=left,
xmajorgrids,
ymajorgrids,
legend style={legend cell align=left, align=left, draw=white!15!black, font=\huge},
x tick label style={font=\huge, yshift=-0.5ex},
y tick label style={font=\huge, xshift=-0.5ex},
label style={font=\Huge}
]
\addplot [color=mycolor1, line width=1.2pt, mark=x, mark size=6.0pt, mark options={solid, mycolor1}]
  table[row sep=crcr]{%
1.359998	0.106909275998925\\
1.440024	0.0972582562991686\\
1.519998	0.156317121899363\\
1.600025	0.162929025940263\\
1.679999	0.156141110081272\\
1.760025	0.151893885280654\\
1.84	0.148464725749708\\
1.920026	0.13788213043699\\
2.000001	0.129628430754835\\
2.080027	0.122102734632653\\
2.16001	0.101897058669135\\
2.240036	0.0848589417521701\\
2.32001	0.068533606457823\\
2.400036	0.0474499965934569\\
2.480011	0.0219287724560147\\
2.560037	0.0116742639880059\\
2.640012	0.0333873518297469\\
2.720038	0.0529897466849647\\
2.800013	0.0719056415525205\\
2.880039	0.101852837779399\\
2.960014	0.202597311518087\\
3.040041	0.240951456949688\\
nan	nan\\
3.52005	0.430959092400732\\
3.600029	0.414650238718776\\
3.680052	0.406441064455004\\
3.760026	0.373322955694183\\
3.840052	0.361062955735758\\
3.920027	0.351679650241689\\
4.000053	0.359885708539224\\
4.080028	0.326248845393001\\
4.160062	0.304348490816144\\
4.240037	0.263132818678834\\
4.320063	0.243206122502096\\
4.400041	0.225615020076644\\
4.480064	0.20738981030043\\
4.560038	0.191325011623745\\
4.640065	0.166816552922048\\
4.720039	0.147909550184412\\
4.800065	0.139337060960669\\
4.88004	0.142468736921311\\
4.960066	0.151840269397301\\
5.040041	0.161264193050814\\
5.120067	0.165343050810504\\
5.20005	0.185908147503422\\
5.280075	0.195982051686711\\
5.36005	0.214091530319793\\
5.440076	0.234477755089108\\
5.520052	0.217146295632055\\
5.600077	0.0989238555919307\\
5.680055	0.0985002725733644\\
5.760078	0.0352003693176002\\
5.840053	0.00928823495390851\\
5.920079	0.00646484255392998\\
6.000054	0.00032884945024314\\
6.08008	0.015658672394073\\
6.160063	0.0181812573769893\\
6.240089	0.0211334374102261\\
6.320063	0.0209460389535536\\
6.40009	0.00996448387466842\\
6.480064	0.000522874377175242\\
6.56009	0.0305430555997699\\
6.640066	0.0548768634528862\\
6.720091	0.0870121291139245\\
6.800066	0.117762195504802\\
6.880092	0.151551621745025\\
6.960067	0.195878438746032\\
7.040096	0.22769163266959\\
7.120068	0.256026851821797\\
7.200102	0.277619791704018\\
7.280077	0.299458235636692\\
7.360103	0.321235700923925\\
7.440077	0.342327789662757\\
7.520103	0.362027561308759\\
7.600078	0.384085467389225\\
7.680107	0.400201840179477\\
7.760079	0.419907746327489\\
7.840105	0.441177234199507\\
7.92008	0.472351786062214\\
8.000107	0.455266256630097\\
8.08008	0.446009613810574\\
8.160115	0.431169677268969\\
8.24009	0.407037153437571\\
8.320116	0.391672844435698\\
8.400092	0.402703343721239\\
nan	nan\\
9.280129	0.0049696378724402\\
nan	nan\\
10.080133	0.0409583589807738\\
10.160118	0.0302968172863327\\
10.240142	0.0913812196150143\\
10.320134	0.132298175920633\\
10.400143	0.204760957121531\\
10.480118	0.249545202058718\\
10.560144	0.281158185668026\\
10.640118	0.306751135512815\\
10.720145	0.327458943712756\\
10.800119	0.357520170310796\\
10.880148	0.386830107149891\\
10.96012	0.410708848278475\\
11.040146	0.430500737863483\\
11.120121	0.447431765650389\\
11.200154	0.455395076309207\\
11.280129	0.470232530286184\\
11.360157	0.475758428636695\\
11.44013	0.476663652228757\\
11.520156	0.46317723674809\\
11.600131	0.449500979886908\\
11.680157	0.431759170301314\\
11.760133	0.271897857024818\\
11.840158	0.332439625085238\\
11.920133	0.334521951002758\\
12.000161	0.335032887937928\\
12.080133	0.338093796932339\\
12.160167	0.357523594894801\\
nan	nan\\
13.520157	0.046585453115874\\
13.600183	0.0505018924902529\\
13.680158	0.0667540140538598\\
13.760184	0.0782949983646297\\
13.840159	0.0829188558218892\\
13.920184	0.0957858865849435\\
14.000159	0.108272258833455\\
14.080186	0.121410985494071\\
14.160168	0.129963651631749\\
14.240197	0.130711223279611\\
14.320169	0.140022753516423\\
14.400195	0.147414948659305\\
14.48017	0.155372749725654\\
14.560196	0.148125430124423\\
14.640171	0.112120073506267\\
14.720197	0.0870317436712202\\
14.800171	0.0747783153487536\\
14.880197	0.0817323749378485\\
14.960172	0.0847281428774096\\
15.040198	0.08447973367646\\
15.120173	0.0827467043687538\\
15.200206	0.0771748994729187\\
15.28018	0.0784213180303777\\
15.360206	0.0743150082302453\\
15.440181	0.0727004350322886\\
15.52021	0.0572062594574252\\
15.600182	0.0472872117438623\\
15.680208	0.0338513132053482\\
15.760183	0.0232213600700621\\
15.840209	0.0178108357347568\\
};
\addlegendentry{Fusion Hypotheses $O_F$}

\addplot [color=mycolor2, dashed, line width=1.2pt, mark=o, mark size=6.0pt, mark options={solid, mycolor2}]
  table[row sep=crcr]{%
1.359998	0.0231599970279177\\
1.440024	0.00600746733668878\\
1.519998	0.361979150539222\\
1.600025	0.238345366022827\\
1.679999	0.194405607845203\\
1.760025	0.266765006570168\\
1.84	0.242327766563458\\
1.920026	0.244061217665971\\
2.000001	0.234335985641176\\
2.080027	0.211793178732485\\
2.16001	0.15656909025045\\
2.240036	0.145270443767137\\
2.32001	0.125592226333623\\
2.400036	0.126849796523299\\
2.480011	0.0523056031314653\\
2.560037	0.0110106330600868\\
2.640012	0.0393804776737721\\
2.720038	0.0703071860403632\\
2.800013	0.0964167185319607\\
2.880039	0.15709752430941\\
2.960014	0.202597311518087\\
3.040041	0.240951456949688\\
nan	nan\\
4.88004	0.208657182102002\\
4.960066	0.187856830408585\\
5.040041	0.194086556573586\\
5.120067	0.184162545953836\\
5.20005	0.202435993722652\\
5.280075	0.194643021696132\\
5.36005	0.213342056720395\\
5.440076	0.215263572154017\\
5.520052	0.217146295632055\\
nan	nan\\
10.080133	0.197104787251925\\
nan	nan\\
13.520157	0.0702871050986928\\
};
\addlegendentry{Object Hypotheses $O_T$}

\end{axis}
\end{tikzpicture}%
		}%
	}
	\caption{Comparison of the results for the position and orientation error in sequence \#1 with both sensor variations for tracking. }
	\label{fig:comparison_seq_1}
\end{figure*}

Thus, there are more time steps with hypotheses from the fusion than from the tracking. To enable now a comparison of position and orientation between the two methods, only the hypotheses are compared if a hypothesis of both methods exists, with a corresponding time tolerance. The differences in the number of measurements can be seen in \cref{fig:comparison_seq_1}.

  \begin{table}
  	\setlength{\tabcolsep}{5pt}
  	\centering
  	\renewcommand{\arraystretch}{1.5}
  	\begin{tabularx}{\columnwidth}{c|c|r|r||r|r||r|r|}
  		\multicolumn{2}{c|}{\multirow{3}{*}{\thead{Sequences\\ ~} }} & \multicolumn{2}{c||}{\thead{RMSE of\\ Position\\ ~[m]}} &  \multicolumn{2}{c||}{\thead{RMSE of\\ Orientation\\ ~[rad]}} 	& \multicolumn{2}{c|}{\thead{Duration of \\Tracking \\ ~[s]}}	\\ \cline{3-8}
  		\multicolumn{2}{c|}{} & V 1		& V 2	& V 1	& V 2	& V 1 	& V 2	\\ \cline{1-8}
        \multirow{3}{*}{\makecell{\thead{\#1}\\ LRR}}  
  		& $O_T$ & 1.12 	& 2.08 	& 0.26 	& 0.21 	& 11.6 	& 23.8	\\ \cline{2-8}
  		& $O_F$ & 0.75 	& 1.63 	& 0.18 	& 0.20 	& 15.2 	& 23.9 	\\ \cline{2-8}
  		& \% 	 & 33.04	& 21.63	& 30.77	& 4.76	& 31.03	& 0.42	\\ \cline{1-8}
  		\multirow{3}{*}{\makecell{\thead{\#1}\\LRR \&\\ibeo}}  
  		& $O_T$ & 1.34 	& 1.10 	& 0.17 	& 0.22 	&  2.3 	& 5.5	\\ \cline{2-8}
  		& $O_F$ & 0.68 	& 0.59 	& 0.13 	& 0.21 	& 11.0 	& 5.9 	\\ \cline{2-8}
  		& \% 	 & 49.25	& 46.36	& 23.53	& 4.55	& 378.26& 7.27\\ \cline{1-8}
  		\multirow{3}{*}{\makecell{\thead{\#2}\\ LRR}}  
  		& $O_T$ & 0.51 	& 3.18 	& 0.02 	& 0.22 	& 3.0 	& 2.7	\\ \cline{2-8}
  		& $O_F$ & 0.20 	& 0.23	& 0.02 	& 0.02 	& 7.9 	& 3.3 	\\ \cline{2-8}
  		& \% 	 & 60.78	& 92.77	& 0.00	& 90.91	& 163.33& 22.22	\\ \cline{1-8}
  		\multirow{3}{*}{\makecell{\thead{\#2}\\ LRR \&\\ibeo}}  
  		& $O_T$ & 0.43 	& 0.76 	& 0.02 	& 0.06 	&  3.0 	& 3.3	\\ \cline{2-8}
  		& $O_F$ & 0.19 	& 0.48 	& 0.02 	& 0.01 	&  7.8 	& 3.3 	\\ \cline{2-8}
  		& \% 	 & 55.81	& 36.84	& 0.00	& 83.33	& 160.00& 0.00	\\ \cline{1-8}
  		\multirow{3}{*}{\makecell{\thead{\#3}\\ LRR}}
  		& $O_T$	 	& 0.54 	& 1.29 	& 0.15 	& 0.21 	& 3.8	& 10.6	\\ \cline{2-8}
  		& $O_F$ 	& 0.40 	& 0.41 	& 0.09 	& 0.11 	& 8.4 	& 13.9 	\\ \cline{2-8}
  		& \% 	 	& 25.93	& 68.22	& 60.00	& 47.62	& 121.05& 31.13 \\ \cline{1-8}
  		\multirow{3}{*}{\makecell{\thead{\#3}\\ LRR \&\\ibeo}}  
  		& $O_T$ 	& 1.16 	& 0.84 	& 0.15 	& 0.09 	&  2.6 	& 9.4	\\ \cline{2-8}
  		& $O_F$ 	& 0.38 	& 0.68 	& 0.09 	& 0.07 	&  5.9 	& 11.9 	\\ \cline{2-8}
  		& \% 	 	& 67.24	& 19.05	& 60.00	& 22.22	& 126.92& 26.60	\\ 
  		\cline{1-8}
  	\end{tabularx}
  	\renewcommand{\arraystretch}{1}
  	\caption{Evaluation table for the sequences with the RMSE for position and orientation as well as the time tracking of objects for vehicles V 1 and V 2. For each sequence a tracking with only LRR and in combination of LRR and ibeo laser scanner was performed. The object hypotheses $O_T$ of the tracking and the hypotheses $O_F$ from the fusion  are compared. In addition, the percentage improvement of the error and the percentage extension for the track time are given in line \%.}
  	\label{tbl:evaluation}
  \end{table}

Up to 3.8 times more was achieved in the duration of object tracking. The graphs of \cref{fig:comparison_seq_1} are shown in detail. In this figures the error due to the euclidean distance to ground truth data of sequence \#1 is plotted. It can be seen that the results of the tracking based on LRR and ibeo data are inferior than with the LRR alone, because there is not enough data available to perform a good box fitting. Tracking with LRR benefits from the long range of the sensor and model assumptions. It is sufficient for the fusion algorithm if an object hypothesis appears temporarily, as is the case for some time step, see \cref{fig:comparison_seq_1_pos_lrr_ibeo,fig:comparison_seq_1_ori_lrr_ibeo} in the tail part. 

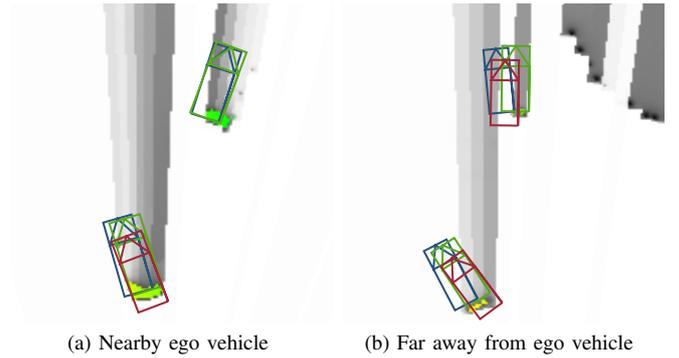
\begin{figure}[th]
	\centering
	\subfloat[Nearby ego vehicle] {
		\label{fig:evaluation_seq_1_schema_nearby}	
		\resizebox{.49\columnwidth}{!}{%
%
\definecolor{mycolor1}{rgb}{0.00000,0.44700,0.74100}%
\definecolor{mycolor2}{rgb}{0.14902,0.32941,0.48627}%
\definecolor{mycolor3}{rgb}{0.33725,0.66667,0.10980}%
\definecolor{mycolor4}{rgb}{0.63922,0.14902,0.21961}%
\begin{tikzpicture}

\begin{axis}[%
width=54.25in,
height=54.25in,
at={(39.161in,1in)},
scale only axis,
point meta min=0,
point meta max=1,
axis on top,
unbounded coords=jump,
every outer x axis line/.append style={none},
every x tick label/.append style={font=\color{none}},
every x tick/.append style={none},
xmin=1,
xmax=136,
xtick={  1,  21,  41,  61,  81, 101, 121},
tick align=outside,
every outer y axis line/.append style={none},
every y tick label/.append style={font=\color{none}},
every y tick/.append style={none},
y dir=reverse,
ymin=1,
ymax=136,
ytick={  1,  21,  41,  61,  81, 101, 121},
axis x line*=bottom,
axis y line*=left,
legend style={legend cell align=left, align=left, draw=white!15!black, fill=white!94!black},
axis lines=none
]
\addplot [forget plot] graphics [xmin=0.5, xmax=137.5, ymin=0.5, ymax=137.5] {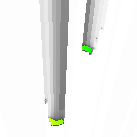};
\addplot[color=mycolor1, point meta={sqrt((\thisrow{u})^2+(\thisrow{v})^2)}, point meta min=0, quiver={u=\thisrow{u}, v=\thisrow{v}, every arrow/.append style={-{Straight Barb[angle'=18.263, scale={10/1000*\pgfplotspointmetatransformed}]}}}]
 table[row sep=crcr] {%
x	y	u	v\\
nan	nan	nan	nan\\
};

\addplot [color=mycolor2, line width=25pt]
  table[row sep=crcr]{%
87.9434636224375	51.1509902174851\\
76.4469915664585	47.485534838008\\
85.9690750992214	17.6201140605421\\
97.4655471552005	21.2855694400191\\
87.9434636224375	51.1509902174851\\
94.5232233435768	30.513984460256\\
91.7173111272109	19.4528417502806\\
83.0267512875977	26.848529080779\\
94.5232233435768	30.513984460256\\
};

\addplot [color=mycolor2, line width=25pt]
  table[row sep=crcr]{%
60.1538619552649	121.607910174005\\
48.2700127669573	125.005514672549\\
39.299530555322	93.6293324844339\\
51.1833797436296	90.23172798589\\
60.1538619552649	121.607910174005\\
53.9552587470249	99.9269682820176\\
45.2414551494758	91.9305302351619\\
42.0714095587173	103.324572780561\\
53.9552587470249	99.9269682820176\\
};

\addplot [color=mycolor3, line width=25pt]
  table[row sep=crcr]{%
88.4480125201271	51.5520867689425\\
75.9463582035768	46.9167439554338\\
87.0711809559976	16.9127735957128\\
99.572835272548	21.5481164092214\\
88.4480125201271	51.5520867689425\\
96.1352650420499	30.8193432503753\\
93.3220081142728	19.2304450024671\\
83.6336107254996	26.1840004368666\\
96.1352650420499	30.8193432503753\\
};

\addplot [color=mycolor3, line width=25pt]
  table[row sep=crcr]{%
64.1728337541995	120.570049899025\\
51.448013117598	124.552097540074\\
41.8910987790784	94.0125280122307\\
54.61591941568	90.030480371181\\
64.1728337541995	120.570049899025\\
57.5690059462826	99.4672073552847\\
48.2535090973793	92.0215041917058\\
44.8441853096811	103.449254996334\\
57.5690059462826	99.4672073552847\\
};

\addplot [color=mycolor4, line width=25pt]
  table[row sep=crcr]{%
nan	nan\\
nan	nan\\
nan	nan\\
nan	nan\\
nan	nan\\
nan	nan\\
nan	nan\\
nan	nan\\
nan	nan\\
};

\addplot [color=mycolor4, line width=25pt]
  table[row sep=crcr]{%
66.5906799698183	126.982178745337\\
54.1270041568351	131.718691056957\\
42.7593746089475	101.805869105797\\
55.2230504219308	97.0693567941772\\
66.5906799698183	126.982178745337\\
58.735647952228	106.312418777086\\
48.9912125154391	99.4376129499871\\
46.2719721392448	111.048931088705\\
58.735647952228	106.312418777086\\
};

\end{axis}

\end{tikzpicture}%
		}%
	}
	\subfloat[Far away from ego vehicle] {
		\label{fig:evaluation_seq_1_schema_far_away}
		\resizebox{.49\columnwidth}{!}{%
%
\definecolor{mycolor1}{rgb}{0.00000,0.44700,0.74100}%
\definecolor{mycolor2}{rgb}{0.14902,0.32941,0.48627}%
\definecolor{mycolor3}{rgb}{0.33725,0.66667,0.10980}%
\definecolor{mycolor4}{rgb}{0.63922,0.14902,0.21961}%
\begin{tikzpicture}

\begin{axis}[%
width=54.25in,
height=54.25in,
at={(39.161in,1in)},
scale only axis,
point meta min=0,
point meta max=1,
axis on top,
every outer x axis line/.append style={none},
every x tick label/.append style={font=\color{none}},
every x tick/.append style={none},
xmin=1,
xmax=157,
xtick={  1,  21,  41,  61,  81, 101, 121, 141},
tick align=outside,
every outer y axis line/.append style={none},
every y tick label/.append style={font=\color{none}},
every y tick/.append style={none},
y dir=reverse,
ymin=1,
ymax=157,
ytick={  1,  21,  41,  61,  81, 101, 121, 141},
axis x line*=bottom,
axis y line*=left,
legend style={legend cell align=left, align=left, draw=white!15!black, fill=white!94!black},
axis lines=none
]
\addplot [forget plot] graphics [xmin=0.5, xmax=158.5, ymin=0.5, ymax=158.5] {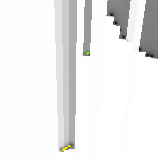};
\addplot[color=mycolor1, point meta={sqrt((\thisrow{u})^2+(\thisrow{v})^2)}, point meta min=0, quiver={u=\thisrow{u}, v=\thisrow{v}, every arrow/.append style={-{Straight Barb[angle'=18.263, scale={10/1000*\pgfplotspointmetatransformed}]}}}]
 table[row sep=crcr] {%
x	y	u	v\\
nan	nan	nan	nan\\
};

\addplot [color=mycolor2, line width=25pt]
  table[row sep=crcr]{%
84.2374337375505	53.6554766177416\\
72.2258185762776	54.8067950316999\\
69.2349350500723	23.6031406182712\\
81.2465502113454	22.451822204313\\
84.2374337375505	53.6554766177416\\
82.1707332209428	32.0937514180625\\
75.2407426307088	23.0274814112921\\
70.1591180596697	33.2450698320206\\
82.1707332209428	32.0937514180625\\
};

\addplot [color=mycolor2, line width=25pt]
  table[row sep=crcr]{%
66.3994723607681	145.228538106299\\
55.4392150024156	150.941886205062\\
40.3546201030372	122.004205639186\\
51.3148774613895	116.290857540422\\
66.3994723607681	145.228538106299\\
55.9760172852975	125.232600835278\\
45.8347487822133	119.147531589804\\
45.0157599269451	130.945948934042\\
55.9760172852975	125.232600835278\\
};

\addplot [color=mycolor3, line width=25pt]
  table[row sep=crcr]{%
91.3593895177412	53.6951345682551\\
78.0269939946958	53.5370419803526\\
78.406416205662	21.5392927250436\\
91.7388117287073	21.6973853129462\\
91.3593895177412	53.6951345682551\\
91.6215702655188	31.5846898328366\\
85.0726139671848	21.6183390189949\\
78.2891747424734	31.4265972449341\\
91.6215702655188	31.5846898328366\\
};

\addplot [color=mycolor3, line width=25pt]
  table[row sep=crcr]{%
72.555282589013	142.340096919125\\
61.3829124024033	149.617177931634\\
43.9179179723803	122.803489483771\\
55.09028815899	115.526408471262\\
72.555282589013	142.340096919125\\
60.4869714378671	123.811838201652\\
49.5041030656852	119.164948977517\\
49.3146012512575	131.088919214161\\
60.4869714378671	123.811838201652\\
};

\addplot [color=mycolor4, line width=25pt]
  table[row sep=crcr]{%
86.0409574278523	60.6047850084315\\
72.7093011459925	60.393350783157\\
73.2167432866515	28.3973757066934\\
86.5483995685114	28.6088099319679\\
86.0409574278523	60.6047850084315\\
86.3915999470477	38.4955662305952\\
79.8825714275815	28.5030928193306\\
73.0599436651879	38.2841320053206\\
86.3915999470477	38.4955662305952\\
};

\addplot [color=mycolor4, line width=25pt]
  table[row sep=crcr]{%
78.3436049227348	147.280757058683\\
67.6423064991565	155.234371061059\\
48.5536328934542	129.551254844471\\
59.2549313170324	121.597640842095\\
78.3436049227348	147.280757058683\\
65.1533314611945	129.533723753021\\
53.9042821052433	125.574447843283\\
54.4520330376162	137.487337755397\\
65.1533314611945	129.533723753021\\
};

\end{axis}

\end{tikzpicture}%
		}%
	}
	\caption{Schematic DOGMa fragment with object hypothesis from tracking (red) with only LRR sensor, fusion (green) and ground truth (blue) from sequence \#1.\vspace{-4ex}}
	\label{fig:evaluation_seq_1_schema}
\end{figure}

For visualization purposes, sections of the DOGMa from sequence \#1 at different distances with object hypotheses from tracking (red), fusion (green) and ground truth data (blue) are shown in \cref{fig:evaluation_seq_1_schema}. For the object tracking only the LRR data are used as input. \cref{fig:evaluation_seq_1_schema} is schematically drawn, so that the overlap of the hypothesis is better recognizable. In the background lies the DOGMa, so that the cells and their amount can be seen as well. In this figures not all grid cells are below the ground truth box, because the DOGMa itself contains inaccuracies. So an error is accumulated by the sensor measurement, the measurement model as well as by the discretization and a small difference due to the processing time. \cref{fig:evaluation_seq_1_schema_nearby} shows a section of the near area, therefore many grid cells are occupied. In this area the object tracking in combination of LRR and ibeo laser scanner is well (see at beginning of \cref{fig:comparison_seq_1}). In contrast, only a few cells are occupied in the far distance (see \cref{fig:evaluation_seq_1_schema_far_away}). The LRR is still provides measurements here, so that hypotheses are also generated by tracking (red). In combination with the ibeo laser data, hypotheses are no longer generated here because the data is too sparse (see at end of \cref{fig:comparison_seq_1}). For this reason, the tracking with combination of LRR and ibeo laser data is worse for this case. This would not create any hypotheses and therefore no red boxes would be displayed. For comparison, this is shown in \cref{fig:comparison_seq_1}. At the beginning there are object hypotheses for both variants and at the end, i.e. in the distance, there are only hypotheses for tracking with LRR.

\begin{figure}[h]
	\centering
	\subfloat[With LRR data] {\label{fig:comparison_seq_2_ori_lrr}
		\resizebox{\columnwidth}{!}{%
			\input{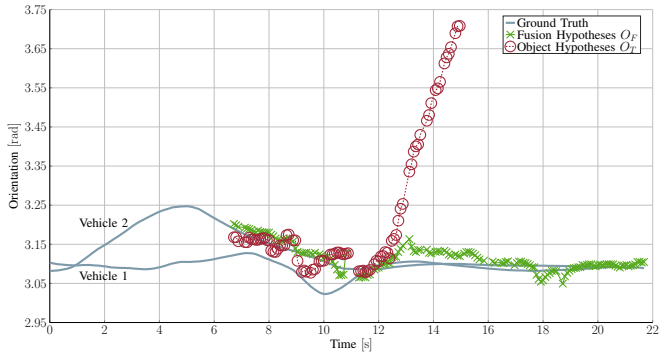}	
		}%
	}\\
	\subfloat[With combination of LRR and ibeo laser scanner data] {\label{fig:comparison_seq_2_ori_lrr+ibeo}
		\resizebox{\columnwidth}{!}{%
			\input{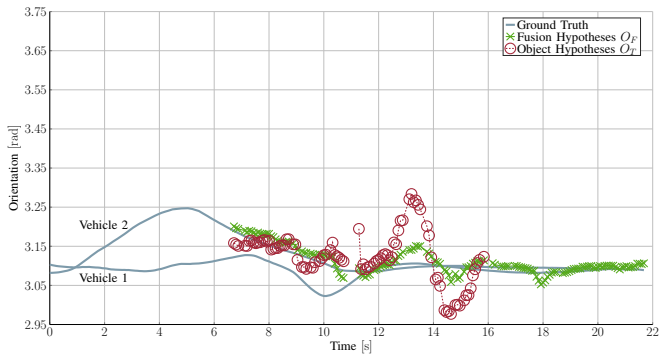}	
		}%
	}
	\caption{Comparison of the orientation evaluation of sequence \#2 with different sensor base for tracking.}
	\label{fig:comparison_seq_2_ori}
\end{figure}

The evaluations result in a reduction of the error in the position by up to \SI{92}{\%} and in the orientation by up to \SI{90}{\%}. This high improvement is due to the fact that only sparse information is available in some object hypotheses and thus position or orientation are not correctly estimated. This applies in particular to sequence \#2. This can be seen in detail for the orientation in \cref{fig:comparison_seq_2_ori}. In contrast to sequence \#1 the tracking with LRR and ibeo laser scanner is now better and keeps the object hypothesis longer and more accurate. 

In tracking, there are situations where only the LRR or the combination with the ibeo laser scanner is better. In both cases, however, the developed fusion achieves significantly improved overall results. In addition, the fusion approach is never inferior to the tracking hypotheses alone.

\section{Conclusion and Outlook}
For environment modeling there exist grid-based and feature-based models, which have different advantages as described above. Therefore, the intentions are to bring both models together to combine their advantages. These two representations are presented for this purpose.

This paper describes a fusion algorithm that combines object hypotheses from tracking with the associated grid cells. In this fusion, the advantage of the object model independent DOGMa is used for a more precise positioning and orientation of the object hypotheses of the LMB tracking \cite{Nuss.2016,Reuter.2014b}. Especially with sparse measurements as well as radar measurements, the object hypothesis of tracking can be well set up and has their benefit.

In the quantitative evaluations, which were carried out on real sequences, it was shown that object hypotheses can be held by the developed fusion algorithm over a longer period of time and the positioning as well as orientation become more accurate. This fusion approach improves the results in general significantly. In addition, any dynamic objects can be tracked since no object assumption was made for the fusion itself. In the implementation, this algorithm is decoupled from the environment models, so that no adjustments are necessary. The fusion was presented with a vehicle tracking, since the developed fusion is independent of object models, it can be used for pedestrians or other dynamic objects.

In future work, various movement models can be used for predicting the previous hypotheses. In addition, an estimate of the extent of objects can be made on the basis of the DOGMa, as suggested in \cite{Gies.2018}.

\section*{Acknowledgment}
In the context of the research of the second author within the thesis to obtain the Bachelor of Engineering, this work is based on that.

In addition, this research is accomplished within the project "UNICARagil" (FKZ 16EMO0290). We acknowledge the financial support for the project by the Federal Ministry of Education and Research of Germany (BMBF).

\addtolength{\textheight}{-12cm}   



\bibliographystyle{IEEEtran}
\bibliography{2019_itsc}

\end{document}